\documentclass[nohyperref]{article}

\usepackage{microtype}
\usepackage{graphicx}
\usepackage{subfigure}
\usepackage{booktabs} 

\usepackage{hyperref}

\usepackage[accepted]{icml2022}

\usepackage{amsmath}
\usepackage{amssymb}
\usepackage{mathtools}
\usepackage{amsthm}
\usepackage{transparent}    
\usepackage{pifont}
\newcommand{\cmark}{\ding{51}}%
\newcommand{\xmark}{\transparent{0.4} \ding{55}}%
\usepackage{physics}

\usepackage[capitalize,noabbrev]{cleveref}

\theoremstyle{plain}

\theoremstyle{definition}

\theoremstyle{remark}

\usepackage[textsize=tiny]{todonotes}

\icmltitlerunning{Neural Laplace: Learning diverse classes of differential equations in the Laplace domain}

\begin{document}

\twocolumn[
\icmltitle{Neural Laplace: Learning diverse classes of differential equations \\ in the Laplace domain}




\begin{icmlauthorlist}
\icmlauthor{Samuel Holt}{yyy}
\icmlauthor{Zhaozhi Qian}{yyy}
\icmlauthor{Mihaela van der Schaar}{yyy}
\end{icmlauthorlist}

\icmlaffiliation{yyy}{Department of Applied Mathematics and Theoretical Physics, University of Cambridge, UK}

\icmlcorrespondingauthor{Samuel Holt}{sih31@cam.ac.uk}

\icmlkeywords{Machine Learning, ICML}

\vskip 0.3in
]



\printAffiliationsAndNotice{}  

\begin{abstract}
Neural Ordinary Differential Equations model dynamical systems with \textit{ODE}s learned by neural networks.
However, ODEs are fundamentally inadequate to model systems with long-range dependencies or discontinuities, which are common in engineering and biological systems. Broader classes of differential equations (DE) have been proposed as remedies, including delay differential equations and integro-differential equations.
Furthermore, Neural ODE suffers from numerical instability when modelling stiff ODEs and ODEs with piecewise forcing functions.
In this work, we propose \textit{Neural Laplace}, a unified framework for learning diverse classes of DEs including all the aforementioned ones.
Instead of modelling the dynamics in the time domain, we model it in the Laplace domain, where the history-dependencies and discontinuities in time can be represented as summations of complex exponentials. To make learning more efficient, we use the geometrical stereographic map of a Riemann sphere to induce more smoothness in the Laplace domain.
In the experiments, Neural Laplace shows superior performance in modelling and extrapolating the trajectories of diverse classes of DEs, including the ones with complex history dependency and abrupt changes.
\end{abstract}

\section{Introduction}
\label{submission}

Learning differential equations that govern dynamical systems is of great practical interest in the natural and social sciences. \citet{DBLP:journals/corr/abs-1806-07366} introduced neural Ordinary Differential Equation (ODE) to model the temporal states $\mathbf{x}(t)$ according to an \textit{ODE} $\dot{\mathbf{x}}=\mathbf{f}(t,\mathbf{x}(t))$, where the function $\mathbf{f}$ is unknown a priori and is learned by a neural network. 

However, there exists much broader classes of DEs, for which Neural ODE cannot model or describe (Table \ref{tab:DEs}). 
These DEs are formulated to capture more general temporal dynamics, which are beyond an ODE's modeling capacity.

For example, the delay differential equation (DDE) and the integro-differential equation (IDE) both include  \textit{historical} states $\mathbf{x}(t - \tau)$, $\forall \tau > 0$; they thus offer a natural way of capturing the impact from history \citep{forde2005delay,Koch2014ModelingOD}. In contrast, ODEs are inadequate to represent such history dependency because they determine the temporal derivative $\dot{\mathbf{x}}$ by the \textit{current} state $\mathbf{x}(t)$ alone.
As a remedy, the users of ODEs often resort to introducing latent variables or additional states, which may not have any semantic meaning or physical interpretation, making the model less transparent \citep{NEURIPS2019_21be9a4b,DBLP:journals/corr/abs-1907-03907}. 

\begin{table}[t!]

	\caption[]{Families of DEs captured by Neural Laplace.} 
	\begin{center}
\begin{tabular}{@{}ll@{}}
\toprule
Model & Equation                                                       \\ \midrule
ODE   & $\dot{\mathbf{x}}=\mathbf{f}(t,\mathbf{x}(t))$                 \\
DDE   & $\dot{\mathbf{x}}=\mathbf{f}(t,\mathbf{x}(t),\mathbf{x}(t-\tau)), \tau\in \mathbb{R}^+$ \\
IDE   & $\dot{\mathbf{x}}=\mathbf{f}(t,\mathbf{x}(t)) + \int_0^t \mathbf{h}(\tau,\mathbf{x}(\tau)) d\tau$                 \\
Forced ODE & $\dot{\mathbf{x}}=\mathbf{f}(t,\mathbf{x}(t),\mathbf{u}(t))$   \\
\midrule
Stiff ODE & $\dot{\mathbf{x}}=\mathbf{f}(t,\mathbf{x}(t)), \exists\ i, j,  \dot{x}_i \gg \dot{x}_j$   \\\bottomrule
\end{tabular}
\end{center}
    \hfill
    \label{tab:DEs}
    \vskip -0.15in
\end{table}

Furthermore, there also exists sub-classes of ODEs that cannot be adequately modeled by Neural ODEs.
Two prime examples are forced ODEs and stiff ODEs (Table \ref{tab:DEs}). 
In the former case, the system dynamics are influenced by an external forcing function $\mathbf{u}(t)$ which may be only \textit{piecewise} continuous in time. In the latter case, the system often involves states operating at different time scales, i.e. $\dot{x}_i \gg \dot{x}_j$ for some $i, j$. 
However, in both cases, the numerical solver employed by Neural ODE would encounter difficulty in accurately solving the initial value problem (IVP, Equation \ref{ODEMain}), which causes problems in training and inference. 
\begin{equation} \label{ODEMain}
\begin{split}
    \mathbf{x}(t) &= \mathbf{x}(0) + \int_{0}^{t} f(\tau,\mathbf{x}(\tau)) d\tau 
     = \text{Solve} (\mathbf{x}(0), f, t)
\end{split}
\end{equation}
\vskip -0.1in

\begin{figure*}[htb]
    \centering
  \includegraphics[width=\textwidth]{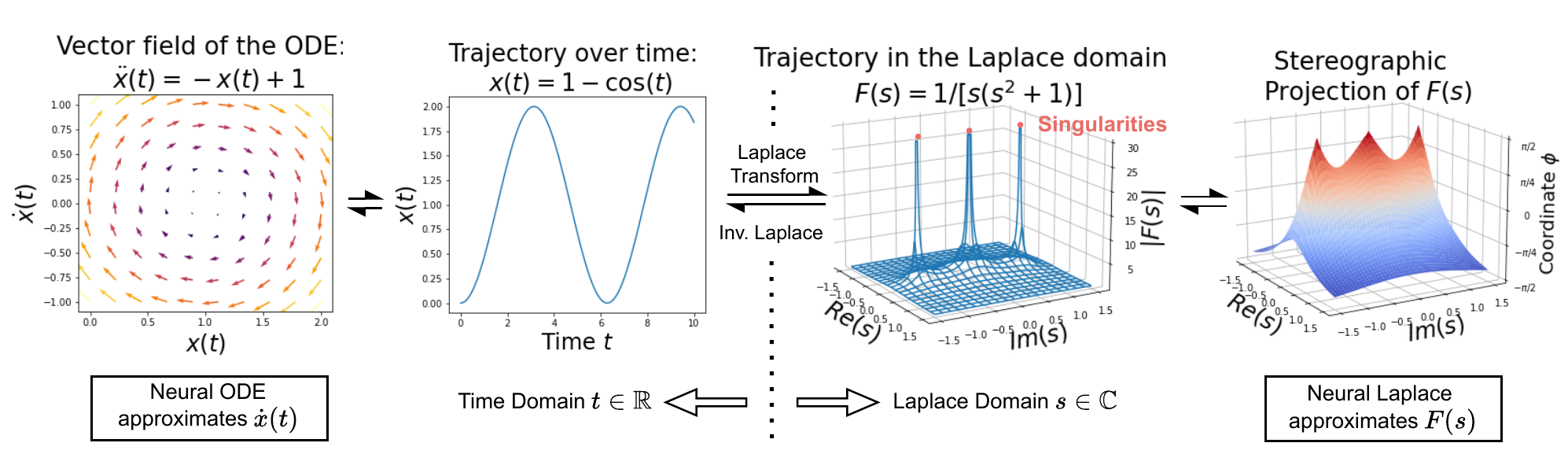}
  \caption{\footnotesize \textbf{Comparison between  Neural Laplace and Neural ODE's modeling approaches}. 
  Neural Laplace models DE solutions in the Laplace domain $F(s)$ and uses the inverse Laplace transform to generate the time solution $x(t)$ at \textit{any} time in the time domain. 
  The Laplace representation $F(s)$ can represent broader classes of DE solutions  than that of an ODE.
  Neural Laplace further uses a stereographic projection to remove the singularities $\infty$ of $F(s)$, forming a continuous and compact domain that improves learning.
  In contrast, Neural ODE models $\dot{x}$ in the time domain using a stepwise numerical ODE solver to generate the time solution. 
  }
  \label{NLModelIllustration}
\end{figure*}

Prior works have attempted to address these limitations separately and individually, which leads to a suite of methods that are often incompatible with each other. 
Whilst there has been some work on adapting Neural ODE to model DDEs specifically \citep{DBLP:journals/corr/abs-2012-06800, DBLP:journals/corr/abs-2102-10801}, their specialized architecture and inference methods rule out the possibility to model other classes of DEs (e.g. IDE or forced ODE). 
Similarly, the works on modeling stiff ODEs have proposed alternative system parameterizations that are invalid for other classes of DEs \citep{DBLP:journals/corr/abs-2110-13040}. 

In this work, we take a holistic approach and propose a unified framework, Neural Laplace, to learn a broad range of DEs (Table \ref{tab:DEs}) which have vast applications (Appendix \ref{deapplications}). Importantly, Neural Laplace does not require the user to specify the class of DE a priori (e.g. choosing between ODE and DDE). Rather with the same network, the appropriate class of DE will be determined implicitly in a data-driven way. 
This significantly extends the flexibility and modeling capability of the existing works.

Unlike the existing works in Neural ODE, which solve the ODE in the time domain (Equation \ref{ODEMain}), Neural Laplace leverages the Laplace transform and models the DE in the Laplace domain (Figure \ref{NLModelIllustration}).
This brings two immediate advantages. 
First, many classes of DEs including DDE and IDE can be easily represented and solved in the Laplace domain \citep{cooke1963differential, yi2006solution, pospisil_representation_2016, cimen2020solution, 10.5555/281875}. 
Secondly, Neural Laplace bypasses the numerical ODE solver and constructs the time solution $\mathbf{x}(t)$ with global summations of complex exponentials (through the inverse Laplace transform, Figure \ref{sdomain}).
It is worth highlighting that the conventional Laplace transform method is used to solve a \textit{known} DE with an analytical form; yet Neural Laplace focuses on learning an \textit{unknown} DE with neural networks and leverages the Laplace transform as a component. 

\textbf{Contributions}. We propose Neural Laplace---a unified framework of learning diverse classes of DEs for modeling dynamical systems. Unlike Neural ODE, Neural Laplace uses neural networks to approximate the DEs in the Laplace domain, which allows it to model general DEs. To facilitate learning and generalization in the Laplace domain, Neural Laplace leverages the stereographic projection of the complex plane on the Reimann sphere. 
Empirically, we show on diverse datasets that Neural Laplace is able to accurately predict DE dynamics with complex history dependencies, abrupt changes, and piecewise external forces, where Neural ODE falls short.

We have released a PyTorch \citep{paszke2017automatic} implementation of Neural Laplace, including GPU implementations of several ILT algorithms. The code for this is at \href{https://github.com/samholt/NeuralLaplace}{https://github.com/samholt/NeuralLaplace}.

\section{Related work}

\begin{table*}[htb]
	
	\caption[]{Comparison with existing works. Neural Laplace is a unified framework of learning diverse classes of DEs. } 
\begin{center}

\begin{tabular}{@{}llccccccc@{}}
\toprule
               &      & Quantity & Initial& \multicolumn{5}{c}{Representable classes of DEs}                             \\ 
Method         & Reference      & Modeled & Condition          &  ODE   & DDE                   & IDE                   & Forced DE              & Stiff DE    \\ \midrule
Neural ODE     & \multicolumn{1}{l|}{\citet{DBLP:journals/corr/abs-1806-07366}} & $\dot{\mathbf{x}}(t)$ &  \multicolumn{1}{c|}{$\mathbf{x}(0)$}  & \cmark & \xmark & \xmark & \xmark & \xmark  \\
ANODE          & \multicolumn{1}{l|}{\citet{NEURIPS2019_21be9a4b}}              & $\dot{\mathbf{x}}(t), \dot{\mathbf{z}}(t)$ &  \multicolumn{1}{c|}{$\mathbf{x}(0), \mathbf{0}$} & \cmark & \xmark & \xmark & \xmark & \xmark \\
Latent ODE     & \multicolumn{1}{l|}{\citet{DBLP:journals/corr/abs-1907-03907}} & $\dot{\mathbf{z}}(t)$ &  \multicolumn{1}{c|}{$\mathbf{z}(0)$} & \cmark & \xmark & \xmark & \xmark & \xmark  \\
ODE2VAE        & \multicolumn{1}{l|}{\citet{yildiz2019ode2vae}}                 & $\dot{\mathbf{x}}(t), \ddot{\mathbf{x}}(t)$ &  \multicolumn{1}{c|}{$\mathbf{x}(0), \dot{\mathbf{x}}(0)$} & \cmark & \xmark & \xmark & \xmark & \xmark  \\
Neural DDE     & \multicolumn{1}{l|}{\citet{DBLP:journals/corr/abs-2102-10801}} & $\dot{\mathbf{x}}(t)$ &  \multicolumn{1}{c|}{$\mathbf{x}(t), t\le 0$}  & \xmark & \cmark & \xmark & \xmark & \xmark\\
Neural Flow    & \multicolumn{1}{l|}{\citet{DBLP:journals/corr/abs-2110-13040}} & $\mathbf{x}(t)$ &  \multicolumn{1}{c|}{$\mathbf{x}(0)$}  & \cmark & \xmark & \xmark & \xmark & \cmark  \\
Neural IM      & \multicolumn{1}{l|}{\citet{gwak2020neural}}                    & $\dot{\mathbf{x}}(t)$ &  \multicolumn{1}{c|}{$\mathbf{x}(0)$}  & \cmark & \xmark & \xmark & \cmark & \xmark \\
Neural Laplace & \multicolumn{1}{l|}{This work}                                 & $\mathbf{F}(\mathbf{s})$&  \multicolumn{1}{c|}{$\mathbf{p}$}  & \cmark & \cmark & \cmark & \cmark & \cmark \\ \bottomrule
\end{tabular}
\end{center}
    \hfill
    \label{tab:related_work}
\end{table*}

Table \ref{tab:related_work} summarizes the key features of the related works and compares them with Neural Laplace. 

Neural ODEs model temporal dynamics with \textit{ODE}s learned by neural networks \citep{DBLP:journals/corr/abs-1806-07366}. 
As a result, Neural ODE inherits the fundamental limitations of ODEs.
Specifically, the temporal dynamics $\dot{\mathbf{x}}(t)$ only depends on the current state $\mathbf{x}(t)$ but not on the history. 
This puts a theoretical limit on the complexity of the trajectories that ODEs can model, and leads to practical consequences (e.g. ODE trajectories cannot intersect). 
Some existing works mitigate this issue by explicitly augmenting the state space \citep{NEURIPS2019_21be9a4b}, introducing latent variables \citep{DBLP:journals/corr/abs-1907-03907} or higher order terms \citep{yildiz2019ode2vae}.
However, they still operate in the ODE framework and cannot model broader classes of DEs. 
Recently, \citet{DBLP:journals/corr/abs-2102-10801} proposes a specialized neural architecture to learn a DDE, but the method is unable to learn the more complex IDE and often suffers from numerical instability. 
\citet{kidger2020neural} proposes to learn history dependency with controlled DEs, but the method requires the trajectory to be twice-differentiable (thus not applicable to systems with abrupt changes). 

Another limitation of Neural ODE and extensions is that they struggle to model certain types of dynamics due to numerical instability.
This is because Neural ODE relies on a numerical ODE solver (Equation \ref{ODEMain}) to predict the trajectory (forward pass) and to compute the network gradients (backward pass). 
Two common scenarios where standard numerical ODE solvers fail are (1) systems with piecewise external forcing or abrupt changes (i.e. discontinuities) and (2) stiff ODEs \citep{DBLP:journals/corr/abs-2006-10711}, both are common in engineering and biological systems \cite{schiff1999laplace}. 
Some existing works address this limitation by using more powerful numerical solvers.
Specifically, when modelling stiff systems, Neural ODE requires special treatment in its computation: either using a very small step size or a specialized IVP numerical solver \citep{kim2021stiff}.
Both lead to a substantial increase in computation cost. However, Neural Laplace does not require special treatment or significant increase in computation for stiff systems.
\citet{DBLP:journals/corr/abs-2110-13040} proposes to model the trajectory $\mathbf{x}(t)$ directly with a neural network, removing the need to use numerical solvers. However, their method cannot model broader classes of DEs or trajectories with abrupt changes. 
\citet{jia2019neural} propose methods that specifically deal with changing external forcing functions, but their proposals are not applicable to other {DDEs and IDEs}.

Furthermore, the Laplace transform has been used to construct input feature filters in the context of scale space theory \citep{lindeberg1996scale, lindeberg2022time}.

\section{Problem and Background}

\textbf{Notation}. For a system with $D\in \mathbb{N}^+$ dimensions, the state of dimension $d$ at time $t$ is denoted as $x_d(t), \forall d= 1, \dots, D, \forall t \in \mathbb{R}$. We elaborate that the \textit{trajectory} $x_d : \mathbb{R} \to \mathbb{R}$ is a function of time, whereas the \textit{state} $x_d(t) \in \mathbb{R}, \forall t \in \mathbb{R}$ is a point on the trajectory. Thus the state vector is $\mathbf{x}(t) := [x_1(t), \dots , x_D(t)]^T \in \mathbb{R}^D$ and the vector-valued trajectory is $\mathbf{x} := [x_1, \dots, x_D]$. The state observations are made at discrete times of $t \in \mathcal{T} = \{t_1, t_2, \dots, T\}$. 

\textbf{Laplace Transform}. The Laplace transform of trajectory $\mathbf{x}$ is defined as \citep{schiff1999laplace}
\begin{equation}
    \mathbf{F}(\mathbf{s})=\mathcal{L}\{\mathbf{x}\}(\mathbf{s})=\int_0^\infty e^{-\mathbf{s}t} \mathbf{x}(t) dt,
\label{mainlt}
\end{equation}
where $\mathbf{s}\in \mathbb{C}^d$ is a vector of \textit{complex} numbers and $\mathbf{F}(\mathbf{s}) \in \mathbb{C}^d$ is called the \textit{Laplace representation}.
The $\mathbf{F}(\mathbf{s})$ may have singularities, i.e. points where $\mathbf{F}(\mathbf{s})\to \mathbf{\infty}$ for one component \citep{schiff1999laplace}.
Importantly, the Laplace transform is well-defined for trajectories that are \textit{piecewise continuous}, i.e. having a finite number of isolated and finite discontinuities \citep{Poularikas2000TheTA}.
This property allows a learned Laplace representation to model a larger class of DE solutions, compared to the {consistently smooth} ODE solutions given by Neural ODE  and variants \citep{NEURIPS2019_21be9a4b}.

\textbf{Inverse Laplace Transform}.
The inverse Laplace transform (ILT) is defined as
\begin{equation}
    \hat{\mathbf{x}}(t) = \mathcal{L}^{-1}\{\mathbf{F}(\mathbf{s})\}(t)=\frac{1}{2\pi i} \int_{\sigma - i \infty}^{\sigma + i \infty} \mathbf{F}(\mathbf{s})e^{\mathbf{s}t}d\mathbf{s},
\label{ilt}
\end{equation}
where the integral refers to the Bromwich contour integral in $\mathbb{C}^d$ with the contour $\sigma>0$ chosen such that all the singularities of $\mathbf{F}(\mathbf{s})$ are to the left of it \citep{schiff1999laplace}. 
Many algorithms have been developed to numerically evaluate Equation \ref{ilt}. On a high level, they involve two steps: \citep{10.1145/321439.321446, deHoog:1982:IMN, kuhlman2012}.
\begin{align}
\label{eq:ILT-Query}
    \mathcal{Q}(t) &= \text{ILT-Query} (t) \\
    \label{eq:ILT-Compute}
    \hat{\mathbf{x}}(t) &= \text{ILT-Compute}\big(\{\mathbf{F}(\mathbf{s})| \mathbf{s} \in \mathcal{Q}(t) \}\big)
\end{align}
To evaluate $\mathbf{x}(t)$ on time points $t \in \mathcal{T} \subset \mathbb{R}$, the algorithms first construct a set of \textit{query points} $\mathbf{s} \in \mathcal{Q}(\mathcal{T}) \subset \mathbb{C}$ (Appendix \ref{ILTaglorithms}). They then compute $\hat{\mathbf{x}}(t)$ using the $\mathbf{F}(\mathbf{s})$ evaluated on these points.
The number of query points scales \textit{linearly} with the number of time points, i.e. $|\mathcal{Q}(\mathcal{T})| = b |\mathcal{T}|$, where the constant $b > 1$, denotes the number of reconstruction terms per time point and is specific to the algorithm. 
Importantly, the computation complexity of ILT only depends on the \textit{number} of time points, but not their values (e.g. ILT for $t=0$ and $t=100$ requires the same amount of computation).  
The vast majority of ILT algorithms are differentiable with respect to $\mathbf{F}(\mathbf{s})$, which allows the gradients to be back propagated through the ILT transform. 
We further discuss the selected ILT in Section \ref{sec:method} and Appendix \ref{ILTaglorithms}.

\begin{figure}[!tb]
      \includegraphics[width=0.95\textwidth/2]{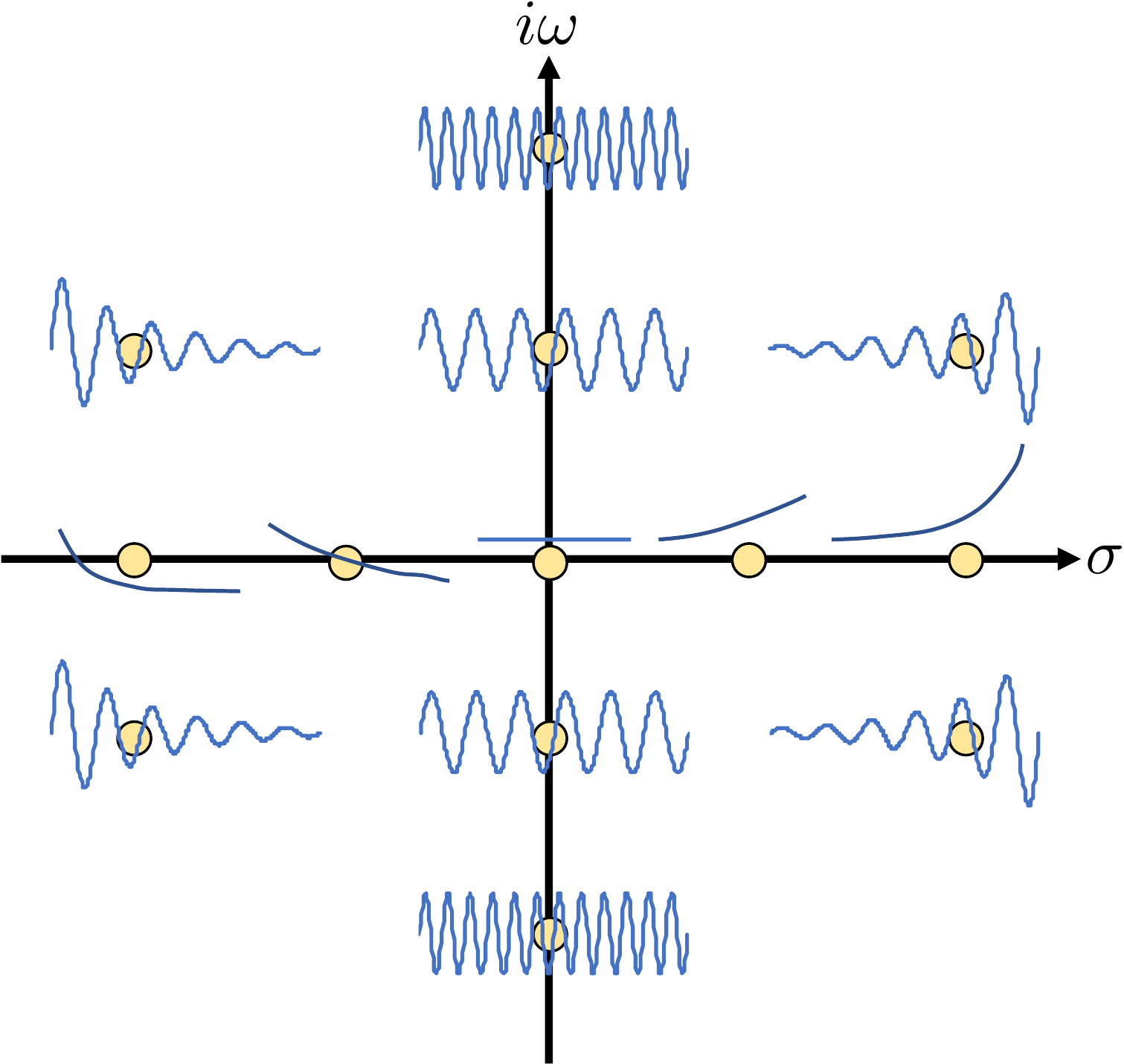}
  \caption{Individual pole points, $s=\sigma + i \omega$ and their time complex exponential reconstructions in blue, illustrated for one dimension of the $s$-domain. $\omega$ is the frequency along the imaginary axis, and $\sigma$ the real component. The complex frequency and its complex conjugate, mean the representation is symmetrical about the real axis $\sigma$. Also the Laplace transform can be seen as a complex generalisation of the Fourier transform, as we get the Fourier transform when $\sigma=0$, i.e. the imaginary axis in our Laplace representation.}
  \label{sdomain}
\end{figure}

Intuitively, the inverse Laplace transform (ILT) (Equation \ref{ilt}) reconstructs the time solution with the basis functions of complex exponentials $e^{\mathbf{s}t}$, which exhibit a mixture of \textit{sinusoidal} and \textit{exponential} components \citep{schiff1999laplace, 10.5555/281875, kuhlman2012}. Figure \ref{sdomain} shows an illustration of these basis function representations.

\begin{figure*}[ht]
    \centering
  \includegraphics[width=0.9\textwidth,trim={0 1cm 0 0},clip]{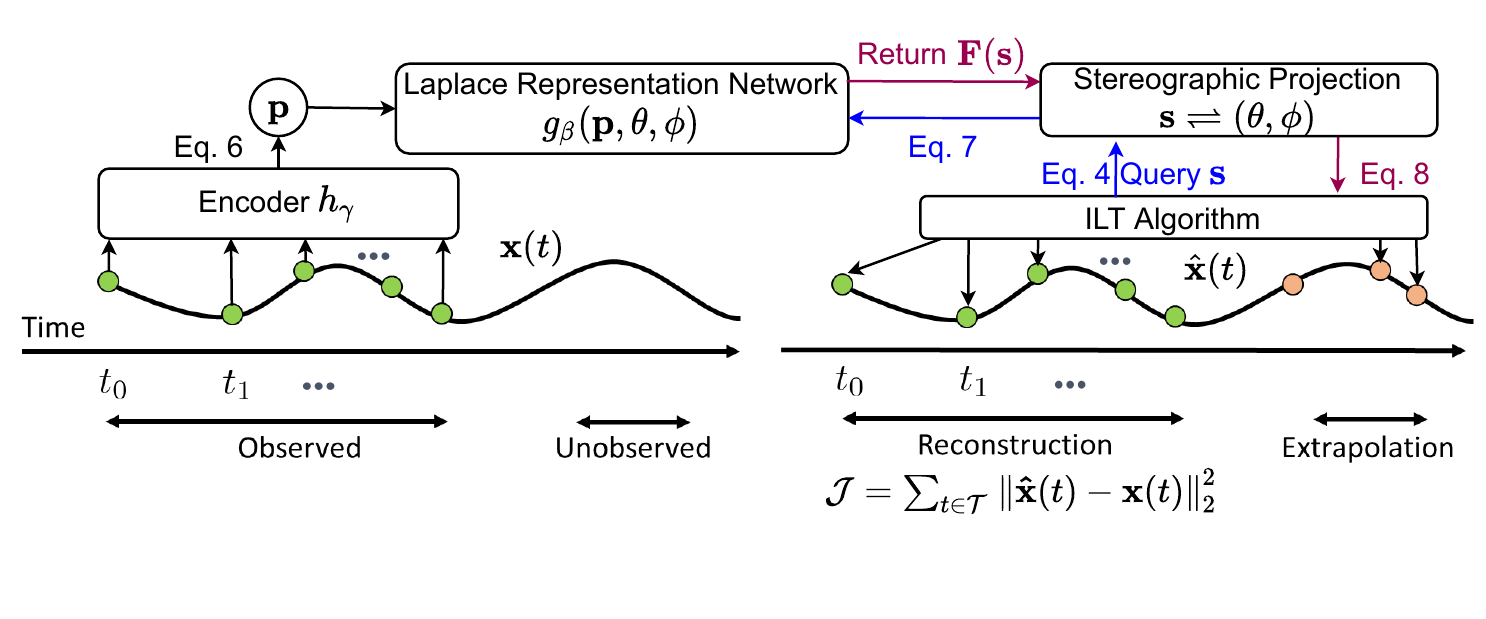}
  \caption{\footnotesize Block diagram of Neural Laplace. The query points $\mathbf{s}$ are given by the ILT algorithm based on the time points to reconstruct or extrapolate. The gradients can be back-propagated through the ILT algorithm and stereographic projection to train networks $h_\gamma$ and $g_\beta$.}
  \label{NLModel}
\end{figure*}

\textbf{Solving DEs in the Laplace domain}. A key application of the Laplace transform is to solve broad classes of DEs, including the ones presented in Table \ref{tab:DEs} \citep{podlubny1997laplace, yousef2018application, yi2006solution, kexue2011laplace}. Due to the Laplace derivative theorem \citep{schiff1999laplace}, the Laplace transform can convert a DE into an \textit{algebraic equation} even when the DE contains historical states $\mathbf{x}(t-\tau)$ (as in DDE), integration terms $\int_0^t \mathbf{h}(\tau,\mathbf{x}(\tau)) d\tau$ (as in IDE) or piecewise continuous terms (as in Forced ODE).
It also applies to coupled DEs and can allow decoupled solutions to coupled DEs for dynamical systems \citep{aastrom2010feedback}.
The resulting algebraic equation can either be solved analytically or numerically to obtain the solution of the DE, $\mathbf{F}(\mathbf{s})$, in the Laplace domain.
Finally, one can obtain the time solution $\mathbf{x}(t)$ by applying the ILT on $\mathbf{F}(\mathbf{s})$. 
As we will show in the next Section, this approach of solving general DEs serves as the foundation of Neural Laplace. 

There also exist numerical simulation techniques in the Laplace domain, the Laplace Transform Boundary Element (LTBE) as a numerical method for solving diffusion-type PDEs \citep{kuhlman2012, Moridis1991TheLT, osti_6901567, Crann2005TheLT} and the Laplace Transform Finite Difference (LTFD) for simulation of single-phase compressible liquid flow through porous media \citep{moridis1991laplace, moridis1994laplace, zahra2017solutions}.

\section{Method}
\label{sec:method}

\textbf{Overview of Neural Laplace.} The Neural Laplace architecture involves three main components: 1. an encoder that learns to infer and represent the initial condition of the trajectory, 2. a Laplace representation network that learns to represent the solutions of DEs in the Laplace domain, and 3. an ILT algorithm that converts the Laplace representation back to the time domain.
The block diagram is shown in Figure \ref{NLModel}. We now discuss each component.

\textbf{Learning to represent initial conditions.} 
The solution of a DE depends on the \textit{initial condition} of the trajectory. Different classes of DEs have different types of initial conditions.
For a first-order ODE (e.g. Neural ODE), it is simply $\mathbf{x}(0)$. For a second-order ODE, it is the vector $(\mathbf{x}(0), \dot{\mathbf{x}}(0))$. And for a DDE with delay $\tau$, it is the function values $\mathbf{x}(t)$, $\forall -\tau \le t\le 0$. 
Note that we only observe the trajectories but do \textit{not} know the class of DE that generates the data. Hence, we need to infer the appropriate initial condition, which implicitly determines the class of DE. 
To achieve this goal, Neural Laplace uses an encoder network to learn a \textit{representation} of the initial condition. We highlight that the observations encoded can be at irregular times.
\begin{equation}
\label{eq:encoder}
    \mathbf{p}=h_\gamma((\mathbf{x}(t_1), t_1), (\mathbf{x}(t_2), t_2), \ldots, (\mathbf{x}(T), T))
\end{equation}
The vector $\mathbf{p} \in \mathbb{R}^K$ is the learned initial condition representation, where $K \ge D$ is a hyper-parameter. The encoder $h$ has trainable weights $\gamma$. Neural Laplace is agnostic to the exact choice of encoder architecture. In the experiments, we use the reverse time gated recurrent unit, similar to \citet{DBLP:journals/corr/abs-1806-07366}, for a fair comparison with the benchmarks. 

\begin{figure}[!htb]
\begin{center}
      \includegraphics[width=0.35\textwidth]{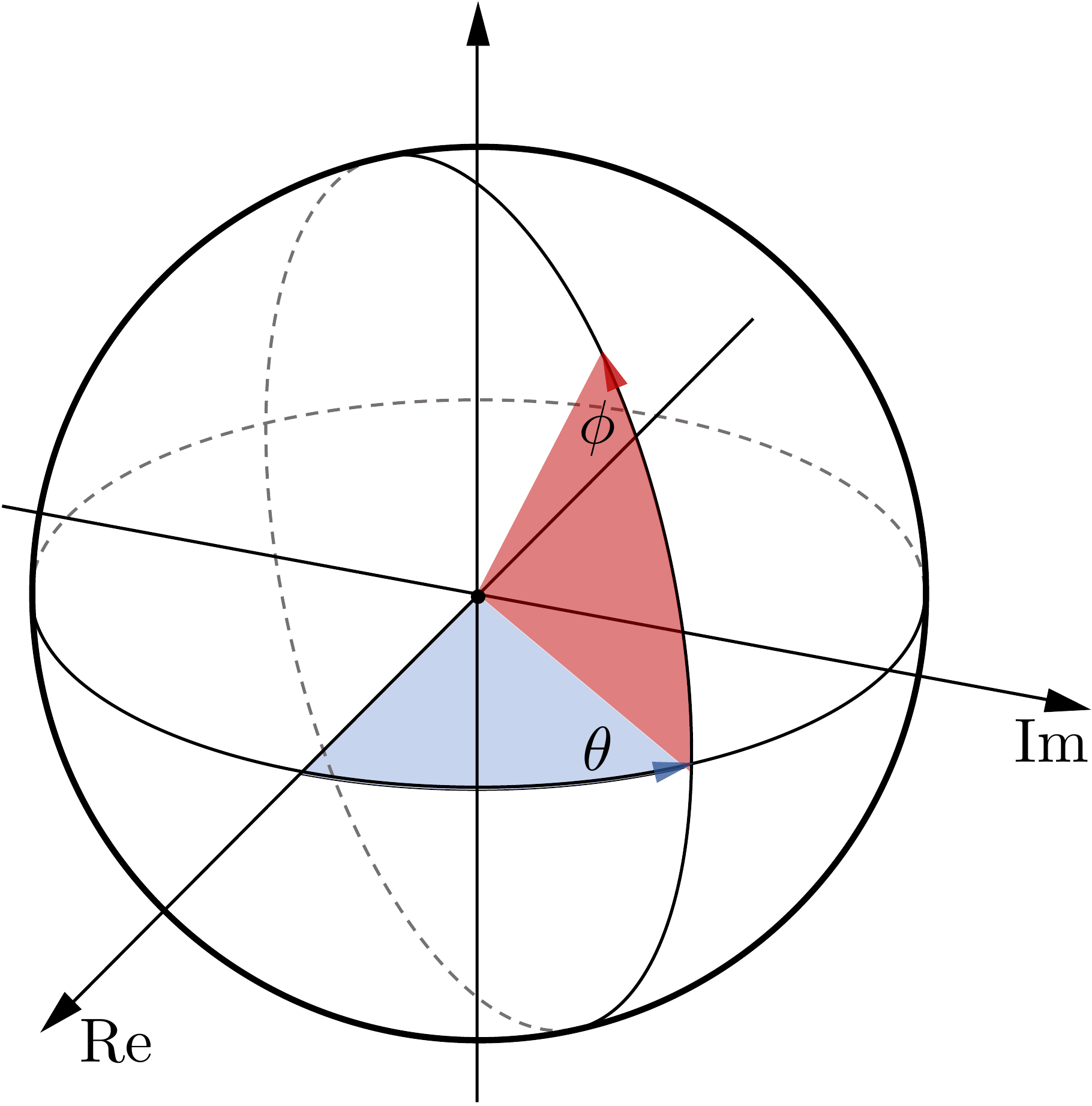}
     \end{center}
  \caption{Geometry of the Riemann sphere map for a complex number $\mathbb{C}$ into a spherical co-ordinate representation of $\theta, \phi$.}
    \label{spheregeom}
\end{figure}

\textbf{Learning DE solutions in the Laplace domain.} 
Given the initial condition representation $\mathbf{p}$, we need to learn a function ${l} : \mathbb{R}^K \times \mathbb{C}^b \to \mathbb{C}^D$ that models the Laplace representation of the DE solution, i.e. $\mathbf{F}(\mathbf{s}) = l(\mathbf{p}, \mathbf{s})$. 
However, the Laplace representation $\mathbf{F}(\mathbf{s})$ often involves singularities \citep{schiff1999laplace}, which are difficult for neural networks to approximate or represent \citep{DBLP:journals/rc/BakerP98}.
We instead propose to use a stereographic projection $(\theta, \phi) = u(s)$ to translate any complex number $s\in \mathbb{C}$ into a coordinate on the Riemann Sphere $(\theta, \phi) \in \mathcal{D} = (-{\pi}, {\pi}) \times (-\frac{\pi}{2}, \frac{\pi}{2})$ \citep{10.5555/26851}, i.e.
\begin{align}
\label{phi0}
     u(s) = \left( \arctan \left( \frac{\Im(s)}{\Re(s)} \right),\arcsin \left( \frac{|s|^2-1}{|s|^2+1} \right) \right)
\end{align}
The inverse transform, $v: \mathcal{D} \rightarrow \mathbb{C}$, is given as
\begin{align}
\label{polarcoords}
    s = v(\theta, \phi) = \tan \left( \frac{\phi}{2} + \frac{\pi}{4} \right) e^{i \theta}
\end{align}
This produces desirable geometrical properties, that a complex point at $\infty$ is the north pole of the sphere $\phi=\frac{\pi}{2}, \forall \theta$ \citep{10.5555/26851}. 
With the stereographic projection, we introduce a feed-forward neural network $g$ to learn the Laplace representation of the DE solution.
\begin{equation}
\label{fs_h}
    \mathbf{F}(\mathbf{s}) =  v \Big(g_\beta \big(\mathbf{p}, u(\mathbf{s})\big) \Big),
\end{equation}
where projections $u$ and $v$ are defined in Equations \ref{phi0} and \ref{polarcoords} respectively, the vector $\mathbf{p}$ is the output of the encoder (Equation \ref{eq:encoder}), and $\beta$ is the trainable weights.
Here the neural network's inputs and outputs are the coordinates on the Riemann Sphere, which is bounded and free from singularities.
Empirically this aids learning and generalization, demonstrating that it can reduce the test RMSE  dramatically compared to learning without the map (Section \ref{ablationstudy}). The improved smoothness with the singularity mapping is shown in Figure \ref{NLModelIllustration}, and geometry in Figure \ref{spheregeom}, for the stereographic projection map of Equation \ref{phi0}.
A nice example of this map is the function of $1/s$, which corresponds to a rotation of the Riemann-sphere $180^{\circ}$ about the real axis. Therefore a representation of $1/s$ under this transformation becomes the map $\theta, \phi \mapsto - \theta, - \phi$ \citep{10.5555/26851}. 

\textbf{Inverse Laplace transform.} After obtaining the Laplace representation $\mathbf{F}(\mathbf{s})$ from Equation \ref{fs_h}, we compute the predicted or reconstructed state values $\hat{\mathbf{x}}(t)$ using the ILT. 
We highlight that we can evaluate $\hat{\mathbf{x}}(t)$ \textit{at any time} $t\in \mathbb{R}$ as the Laplace representation is independent of time once learnt.
In practice, we use the well-known ILT Fourier series inverse algorithm (ILT-FSI), which can obtain the most general time solutions whilst remaining numerically stable \citep{10.1145/321439.321446, deHoog:1982:IMN, kuhlman2012}.
In Appendix \ref{ILTaglorithms}, we provide more details of the ILT-FSI and comparisons with other ILT algorithms.

We note that numerically estimating the LT on the observations $x(t)$ only gives  $\mathbf{F}(\mathbf{s})$ on a \textit{finite} set, $\forall \mathbf{s} \in \mathcal{U} \subset \mathbb{C}$, where $\mathcal{U}$ is determined by the observation times.
Thus, this cannot \textit{generalize} for extrapolation or interpolation. Whereas Neural Laplace learns $\mathbf{F}(\mathbf{s}), \forall \mathbf{s} \in \mathbb{C}$ on the \textit{entire} complex domain $\mathbb{C}$.

\textbf{Loss function.} 
Neural Laplace is trained end-to-end using the mean squared error loss,
\begin{equation}\label{lossfunc}
    \mathcal{J} =  \sum_{t\in \mathcal{T}}  \left\| \mathbf{\hat{x}}(t) - \mathbf{x}(t) \right\|_2^2,
\end{equation}
where $\mathbf{\hat{x}}(t)$ is the reconstructed trajectory (Equation \ref{eq:ILT-Compute}).
We minimize the above loss function $\mathcal{J}$ to learn the encoder $h_\gamma$ and the Laplace representation network $g_\beta$. This training is summarized in Algorithm \ref{alg:training}.
We can also constrain the ILT reconstruction frequencies with a low pass filter, smoothing the reconstructed signal, and we empirically show a toy noise removal task of this in Appendix \ref{frequencyfilter}.

\begin{algorithm}[H]
\caption{Neural Laplace Training Procedure} \label{alg:training}
\begin{algorithmic}
\STATE {\bfseries Input:} Observed trajectory $\mathbf{x}(t),  t \in \mathcal{T}$
\STATE {\bfseries Input:} Training iterations $I$
\FOR{$i \in 1:I$}
\STATE $\mathbf{p} = h_\gamma\big(\{\mathbf{x}(t)| t \in \mathcal{T}\}\big)$  \hfill Eq. \ref{eq:encoder}
\FOR{$t \in \mathcal{T}$}
\STATE  $\mathcal{Q} = \text{ILT-Query} (t)$  \hfill Eq. \ref{eq:ILT-Query}
\FOR{$s \in \mathcal{Q}$}
\STATE $\mathbf{F}(\mathbf{s}) =  v \Big(g_\beta \big(\mathbf{p}, u(\mathbf{s})\big) \Big)$ \hfill Eq. \ref{fs_h}
\ENDFOR
\STATE $\hat{\mathbf{x}}(t) = \text{ILT-Compute}\big(\{\mathbf{F}(\mathbf{s})| \mathbf{s} \in \mathcal{Q} \}\big)$ \hfill Eq. \ref{eq:ILT-Compute}
\ENDFOR
\STATE $\mathcal{J} =  \sum_{t\in \mathcal{T}}  \left\| \mathbf{\hat{x}}(t) - \mathbf{x}(t) \right\|_2^2$ \hfill Eq. \ref{lossfunc}
\STATE Compute gradients of $\mathcal{J}$ via back propagation.
\STATE Update neural network weights $\gamma$ and $\beta$.
\ENDFOR
\STATE {\bfseries Output:} the trained neural networks $h_\gamma$ and $g_\beta$
\end{algorithmic}
\end{algorithm}

\begin{table*}[htb]
	\centering
	\caption[]{Each DE system we use for comparison against the benchmarks, and their properties for comparison.} 
	  \resizebox{\textwidth}{!}{
		\begin{tabular}[b]{l c c c c c c c}
			\toprule
System                              & Piecewise & Discontinuous & Integro & Delay & Stiff & Periodic  \\
                                    & DE        & differential  & DE      & DE    & solutions & solutions \\ \midrule
Spiral DDE                          & \cmark    & \cmark & \xmark & \cmark & \xmark & \xmark \\
Lotka-Volterra DDE                  & \cmark    & \cmark & \xmark & \cmark & \xmark & \xmark \\
Mackey–Glass DDE                    & \cmark    & \cmark & \xmark & \cmark & \xmark & \xmark \\
Stiff Van der Pol Oscillator DE     & \xmark    & \cmark & \xmark & \xmark & \cmark & \cmark \\
ODE with piecewise forcing function & \cmark    & \xmark & \xmark & \xmark & \xmark & \xmark \\
Integro DE                          & \xmark    & \xmark & \cmark & \xmark & \xmark & \xmark \\
                     \bottomrule
        \end{tabular} 
        }
    \hfill
    \vskip -0.15in
    \label{datasetprop}
\end{table*}

\textbf{Comparison with Neural ODE.}
Here we articulate the three main differences between Neural Laplace and Neural ODE. 
\textbf{(1)} The encoders of these two frameworks serves different purposes. The Neural ODE encoder is tasked to infer the initial condition $\mathbf{x}(0)$ when it is observed with noise or unobserved (e.g. measurement starts at $t>0$). On the other hand, the Neural Laplace encoder needs to learn an appropriate representation of the initial condition and implicitly decide the class of DE to use. The representation $\mathbf{p}$ may include more information than $\mathbf{x}(0)$. 
\textbf{(2)} Neural ODE uses a neural network to approximate $\dot{\mathbf{x}}(t)$ while Neural Laplace uses a neural network to approximate $\mathbf{F}(s)$ after the stereographic projection. As a result, Neural ODE can only model \textit{twice-differentiable} trajectories $\mathbf{x}(t)$ while Neural Laplace can model non-smooth trajectories. 
\textbf{(3)} Neural ODE uses numerical IVP solvers while Neural Laplace uses ILT algorithms. The ILT algorithms can handle stiff ODEs and piecewise forcing functions, where most numerical IVP solvers fail \citep{DBLP:journals/corr/abs-2110-13040}. Furthermore, the time complexity of ILT for predicting $\mathbf{x}(t)$ does \textit{not} depend on $t$, while numerical IVP solvers do. This brings computational benefits to Neural Laplace when the application involves a long time horizon. 

\section{Experiments} \label{experiments}

We evaluate Neural Laplace on a broad range of dynamical systems arising from engineering and natural sciences. These systems are governed by different classes of DEs. We show that Neural Laplace is able to model and predict these systems better than the ODE based methods.

\textbf{Benchmarks}. We compare against the standard and augmented Neural ODE (NODE, and ANODE respectively) with an input fixed initial condition \citep{NEURIPS2019_21be9a4b, DBLP:journals/corr/abs-1806-07366}.
We also compare with ODE models with an encoder-decoder architecture: Latent ODE with an ODE-RNN encoder \citep{DBLP:journals/corr/abs-1907-03907}, Neural Flows (NF) Coupling, and Neural Flows ResNet \citep{DBLP:journals/corr/abs-2110-13040}. 
To ensure a fair comparison, we set the number of hidden units per layer such that all models have roughly the same number of total parameters. Further details of hyperparameters and implementation details are in Appendix \ref{benchmarkdetails}. 

\textbf{Evaluation}. 
To test whether the models are able to accurately uncover the temporal dynamics, we evaluate their accuracy in \textit{predicting the future states} of the system, i.e. the root mean square error (\textbf{RMSE}). 
We also evaluate the model's ability to \textit{capture the state space distribution} $P(\mathbf{x}(t))$ by calculating the conditional mutual information (\textbf{CMI}) between the true and the predicted distributions conditioning on the initial value distribution \footnote{The state space distribution portrays many key properties of the dynamical system such as the attractor geometry. It thus has been routinely examined in the literature \citep{schmidt2020identifying}.}. 
We split each sampled trajectory into two equal parts $[0,\frac{T}{2}]$, $[\frac{T}{2}, T]$, encoding the first half and predicting the second half. For each dataset equation we sample $1,000$ trajectories of $200$ time points in the interval of $t\in[0,20]$, with each sampled from a different initial condition giving rise to a unique trajectory defined by the same differential equation system. 
We divide the trajectories into a train-validation-test split of $80:10:10$, for training, hyperparameter tuning, and evaluation respectively. 
See Appendix \ref{samplingdes} for details on how we sampled each DE system.

\textbf{Dynamical systems for comparison}. We selected a broad range of dynamical systems from applied sciences, and each have unique properties of interest, see Table \ref{datasetprop} for a comparison. 
The systems are detailed as follows.

\textbf{Spiral DDE}, \citep{DBLP:journals/corr/abs-2102-10801} these are common in healthcare and biological systems, for example cardiac tissue models \citep{moreira2019delay}, biological networks \citep{glass2021nonlinear} and modelling gene dynamics \citep{verdugo2007delay}.
\begin{equation} \label{spiraldde}
\begin{split}
    \mathbf{\dot{x}}(t) = \mathbf{A} \tanh(\mathbf{x}(t) + \mathbf{x}(t-\tau)) , \quad t > 0 \\
\end{split}
\end{equation}
with the time delay $\tau=2.5$ and $\mathbf{A}\in \mathbb{R}^{2\times2}$ a constant matrix. We generate trajectories by sampling from a grid for each state dimension of the fixed initial history $\mathbf{x}(t), t\leq0$ in the interval $[-2, 2]$.
\begin{table*}[htb]
	\centering
	\caption[]{Test RMSE for datasets analyzed. Best results bolded. Averaged over $5$ runs.} \label{Table:MainMSE}
	  \resizebox{\textwidth}{!}{
		\begin{tabular}[b]{c c c c c c c}
			\toprule
                        & Spiral                       & Lotka-Volterra              & Mackey-Glass                    & Stiff Van der                   & ODE piecewise                 & Integro                          \\
Method                  & DDE                          & DDE                         &  DDE                            & Pol Oscillator DE               &  forcing function             & DE                               \\ \midrule
NODE                    &  .0389 $\pm$ .0029           & .3102 $\pm$ .0151           &  .8225 $\pm$ .0403              & .2833 $\pm$ .0032               & .2274 $\pm$ .0298             & .0730 $\pm$ .0016                \\
ANODE                   & .0365 $\pm$ .0011            & .2930 $\pm$ .0239           &  .8214 $\pm$ .0415              & .2444 $\pm$ .0167               & .0644 $\pm$ .0211             & .0036 $\pm$ .0003                \\
Latent ODE              & .0481 $\pm$ .0033            & .2182 $\pm$ .0153           &  \textbf{.0385 $\pm$ .0217}     & .1932 $\pm$ .0154               & .1401 $\pm$ .0457             & .0109 $\pm$ .0009                \\
NF Coupling             & .6938 $\pm$ .1036            & .7266 $\pm$ .0310           &  \textbf{.0539 $\pm$ .0181}     & .1829 $\pm$ .0209               & .0752 $\pm$ .0052             & .0042 $\pm$ .0013                \\
NF ResNet               & .1905 $\pm$ .0479            & .2257 $\pm$ .0608           &  \textbf{.0350 $\pm$ .0223}     & \textbf{.1468 $\pm$ .0396}      & .0399 $\pm$ .0119             & .0027 $\pm$ .0004                \\
Neural Laplace          & \textbf{.0331	$\pm$ .0023}   & \textbf{.0475 $\pm$ .0061}  & \textbf{.0282 $\pm$ .0246}      & \textbf{.1314 $\pm$ .0218}      & \textbf{.0035 $\pm$ .0004}	 & \textbf{.0014 $\pm$ .0005}	    \\
                     \bottomrule
        \end{tabular} 
        }
    \hfill
    \vskip -0.15in
\end{table*}

\textbf{Lotka-Volterra DDE} \citep{BAHAR2004364}, also known as the predator-prey equations, are fundamental to ecology and  population modeling.
\begin{equation} \label{lotkavolterradde}
    \dot{x} = \frac{1}{2} x(t)(1-y(t-\tau)); \quad
    \dot{y} = \frac{1}{2} y(t)(1-x(t-\tau))
\end{equation}

We use a fixed delay of $\tau=0.1$, generating trajectories by sampling from a grid for each state dimension of the fixed initial history $\mathbf{x}(t), t\leq0$ in the interval $[0.1, 2]$, and instead sample time points in the interval of $t \in [0.1, 2]$.

\begin{figure}[!ht]
\centering
      \includegraphics[width=0.95\textwidth/2]{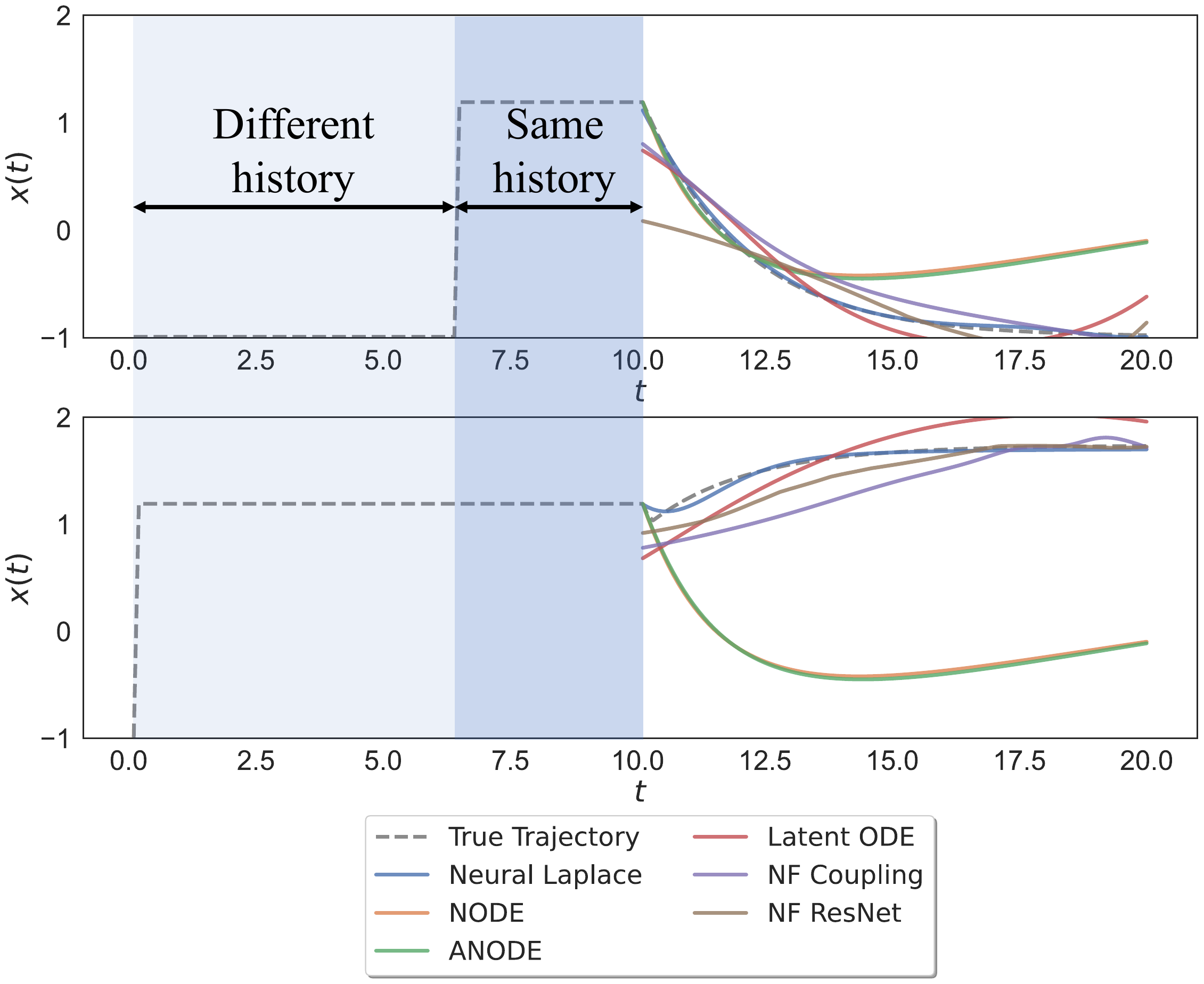}
  \caption{\footnotesize Two test trajectories of the Mackey Glass DDE, with benchmarks, illustrating examples where the distant history impact the future trajectory, even though momentarily they may have the same history for short times, $6 \leq t \leq 10$.}
  \label{mackeylongrange}
\end{figure}

\textbf{Mackey–Glass DDE}, \citep{doi:10.1126/science.267326}, modified to exhibit long range dependencies, given the form,
\begin{equation} \label{stiffvdp_0}
    \dot{x} = \beta \frac{x(t-\tau)}{1+x(t-\tau)^n} - \gamma x(t)
\end{equation}
Using a fixed delay of $\tau=10, n=10, \beta=0.25, \gamma=0.1$, generating trajectories by uniformly changing the initial history to be either $-1$, or $1.1$ over the time interval of $[0,10]$ for $t<10$, as seen in Figure \ref{mackeylongrange}.

\textbf{Stiff Van der Pol Oscillator DE}, \citep{van1927frequency}, which exhibits regions of high stiffness when setting $\mu=$1,000,
\begin{equation} \label{stiffvdp_1}
    \dot{x} = y; \quad
    \dot{y} = \mu (1- x^2) y - x 
\end{equation}
We sample initial conditions from $x(0) \in [0.1, 2], y(0)=0$.

\textbf{ODE with piecewise forcing function}, an ODE with a ramp loading forcing function (common in engineering applications) \citep{boyce2017elementary}. This exhibits piecewise DE behaviour, i.e. a different ODE in the different forcing function piecewise regions.
\begin{equation} \label{ode_discontinous_ff}
    \ddot{x} + 4 x(t) = u(t); \quad
    u(t) = \begin{cases}
       0 &\  0<t \leq 5\\
       \frac{t-5}{5} &\  5<t \leq 10\\
       1 & \  10<t< 20 \\
     \end{cases} \\
\end{equation}
We sample initial conditions from $x(0) \in [0, 0.1], \dot{x}(0)=0$.

\textbf{Integro DE}, Integral and differential system \citep{integrode}.
\begin{equation} \label{integro_de}
\begin{split}
    \dot{x} + 2x(t) + 5 \int_0^t x(t) dt & = u(t) \\
\end{split}
\end{equation}
Where $u(t)$ is the Heaviside step function, sampling initial conditions from $x(0) \in [0, 1]$ and sampling times in the interval of $t \in [0, 4]$.

In Appendix \ref{toywaveforms}, we compare our method on periodic waveforms that are \textit{not} governed by a standard DE. We observe Neural Laplace is better at reconstruction compared to the benchmarks in extrapolating these non-DE trajectories. 

\subsection{Results Discussion}

The RMSE for each dataset comparing against the benchmarks, are tabulated in Table \ref{Table:MainMSE}.
Neural Laplace achieves low RMSE extrapolation test error on all DE datasets analyzed. 
The CMI metric follows a similar pattern and is presented in Appendix \ref{capturingstatespacedistribution}. We also observe a similar pattern on DE datasets corrupted with noise in Appendix \ref{dopri5results} and on DE datasets with smaller sizes, of trajectories and observations in Appendix \ref{samplesize}.
Neural Laplace is able to correctly \textit{learn} the DE  system using its global complex exponential basis function representation through encoding the observed trajectory into the initial condition representation for that DE system, and extrapolating forwards in time. We analyse a few typical scenarios in detail to gain a better understanding. See Appendix \ref{datasetplots} for additional analysis and visualization.

\textbf{Systems with long range dependency}. 
The experiments on Mackey–Glass DDE offers insight into the model's ability to capture long range dependencies. 
As illustrated in Figure \ref{mackeylongrange}, trajectories with the same recent history ($6\le t \le 10$) but different distant histories ($0\le t < 6$) evolve very differently in the future ($t > 10$). Hence, successful extrapolation requires the model to keep memory from the distant past. 
NODE and ANODE methods fail to capture this because $x(10)$, the initial condition for the ODE to extrapolate, is the same for both trajectories---this leads to the same extrapolation (Figure \ref{mackeylongrange}). 
Latent ODE and Neural Flow methods start to encode some history dependency. However they still fail to capture the true solution, being overly smooth and unable to capture the piecewise initial history. Whereas Neural Laplace is able to correctly learn the historical dependency of the DE system. Similar patterns are observed in other systems with long range dependencies (e.g. Spiral and Lotka-Volteera DDE) and further illustrated in Appendix \ref{datasetplots}.
\begin{table*}[!t]
	\centering
	\caption[]{Neural Laplace ablation study and sensitivity analysis. RMSE is reported under different study configurations on different dynamical systems.  Averaged over $5$ runs.} \label{Table:ablate}
		\begin{tabular}[b]{c c c c c c}
			\toprule
                &                   & Lotka-Volterra              & Stiff Van der                & ODE piecewise                & Integro                                  \\
Study           &     Config.       & DDE                         & Pol Oscillator DE            & forcing function             & DE                                       \\ \midrule
Stereographic   & \xmark          	& .1617 $\pm$ .0741           & .1836 $\pm$ .0586            & .0249 $\pm$ .0066	        &	.0048	$\pm$ .0007	                   \\
projection      & \cmark  	        & \textbf{.0614 $\pm$ .0469}  & \textbf{.1286 $\pm$ .0170}   & \textbf{.0036 $\pm$ .0007}   &	\textbf{.0013	$\pm$ .0003}		   \\ \midrule
                & 1             	& .4416 $\pm$ .0898           & .1520 $\pm$ .0240            & .0036 $\pm$ .0007	        &	.0010	$\pm$ .0002	                   \\
                & 2             	& .0405 $\pm$ .0113           & .1308 $\pm$ .0159            & .0033 $\pm$ .0009	        &	.0012	$\pm$ .0002	                   \\
Dimensionality  & 4             	& .0427 $\pm$ .0049           & .1334 $\pm$ .0103            & .0038 $\pm$ .0013	        &	.0012	$\pm$ .0003	                   \\
$K$             & 8             	& .0408 $\pm$ .0134           & .1294 $\pm$ .0173            & .0038 $\pm$ .0004	        &	.0010	$\pm$ .0002	                   \\
                & 16            	& .0380 $\pm$ .0053           & .1334 $\pm$ .0197            & .0036 $\pm$ .0006	        &	.0012	$\pm$ .0005	                   \\
                & 32             	& .0398 $\pm$ .0045           & .1337 $\pm$ .0203            & .0032 $\pm$ .0005	        &	.0013	$\pm$ .0003	                   \\
                     \bottomrule
        \end{tabular} 
    \hfill
\end{table*}
\begin{figure}[!t]
    \centering
    \includegraphics[width=\columnwidth]{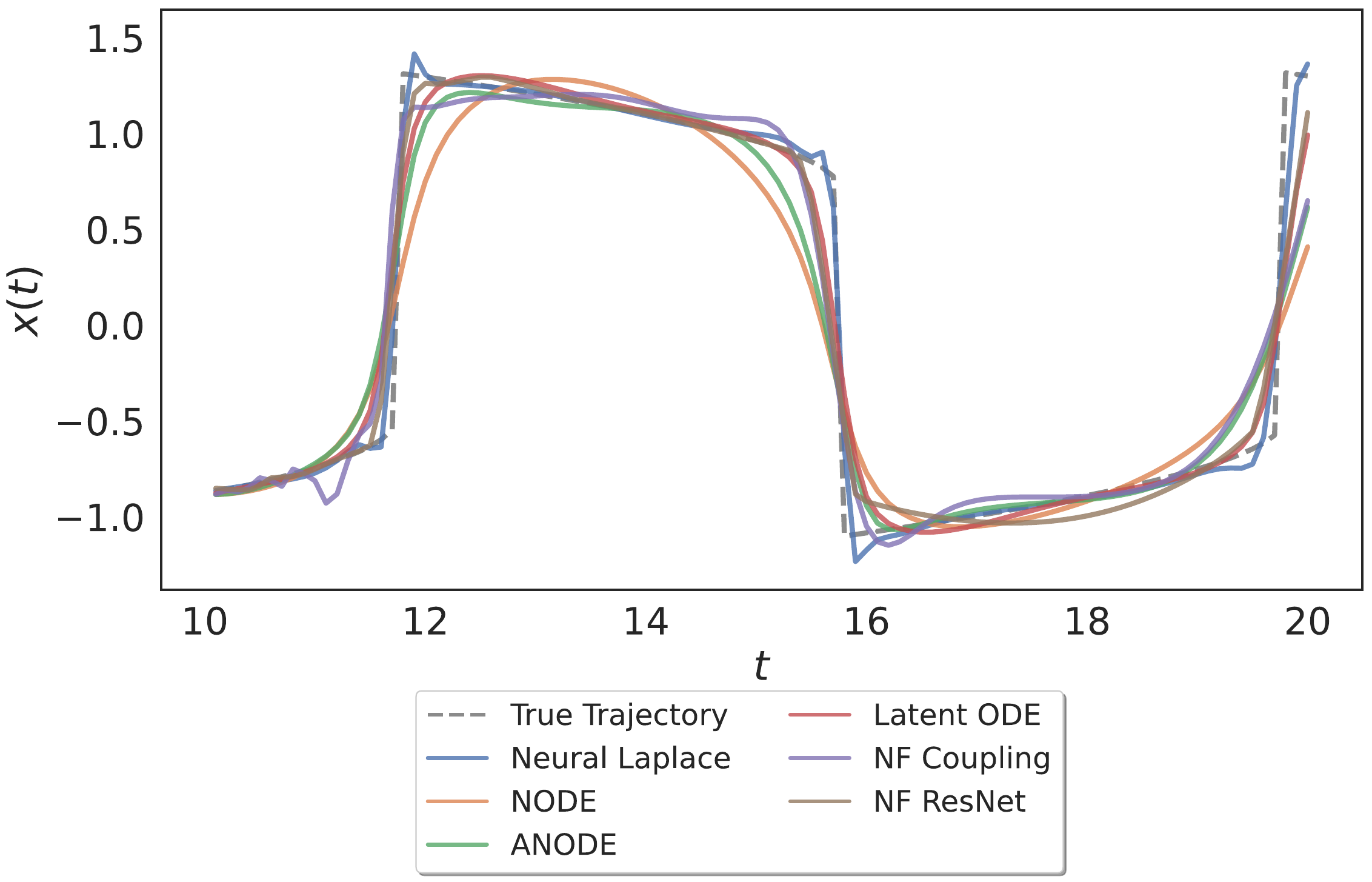}
  \caption{\footnotesize Test extrapolation plots with benchmarks for Stiff Van der Pol Oscillator DE. Neural Laplace is able to correctly extrapolate the DE dynamics.}
  \label{joint_trajectories_plot}
  \vskip -0.15in
\end{figure}

\textbf{Systems with abrupt changes and stiffness.} 
The trajectory plot in Figure \ref{joint_trajectories_plot}, of a test trajectory, shows that Neural Laplace is able to correctly learn the periodic stiff solution, capturing the discontinuities of the derivative of the solution and the periodicity. NODE, ANODE and Latent ODE methods, correctly capture the periodicity, however fall short in modelling the derivative discontinuities as they enforce overly smooth solutions. Neural Flow methods suffer from similar smoothness behaviour.

\subsection{Ablation study and sensitivity analysis} \label{ablationstudy}

\textbf{Ablation study for stereographic projection}.
We investigate how useful the stereographic projection is for learning a Laplace representation of a DE system. Table \ref{Table:ablate} (top) shows the test RMSE with and without it in Neural Laplace. This demonstrates empirically that using the stereographic projection Riemann sphere map can allow us to achieve an order of magnitude reduced test RMSE. This supports our belief that the stereographic projection improves learning by inducing a more compact geometry in the Laplace domain.

\textbf{Sensitivity to dimensionality $K$}. In Neural Laplace, the encoded representation of the initial condition $\mathbf{p} \in \mathbb{R}^K$ has dimensionality $K$, a hyperparameter. We explore the sensitivity of this dimension, reporting the test RMSE in Table \ref{Table:ablate}. This empirically shows that the performance is not sensitive to the exact choice of $K$, as long as it is set to a large enough value (e.g. $K\geq2$).

\subsection{Computation time and complexity}

\begin{figure}[!tb]
  \centering
  \includegraphics[width=0.96\textwidth/2]{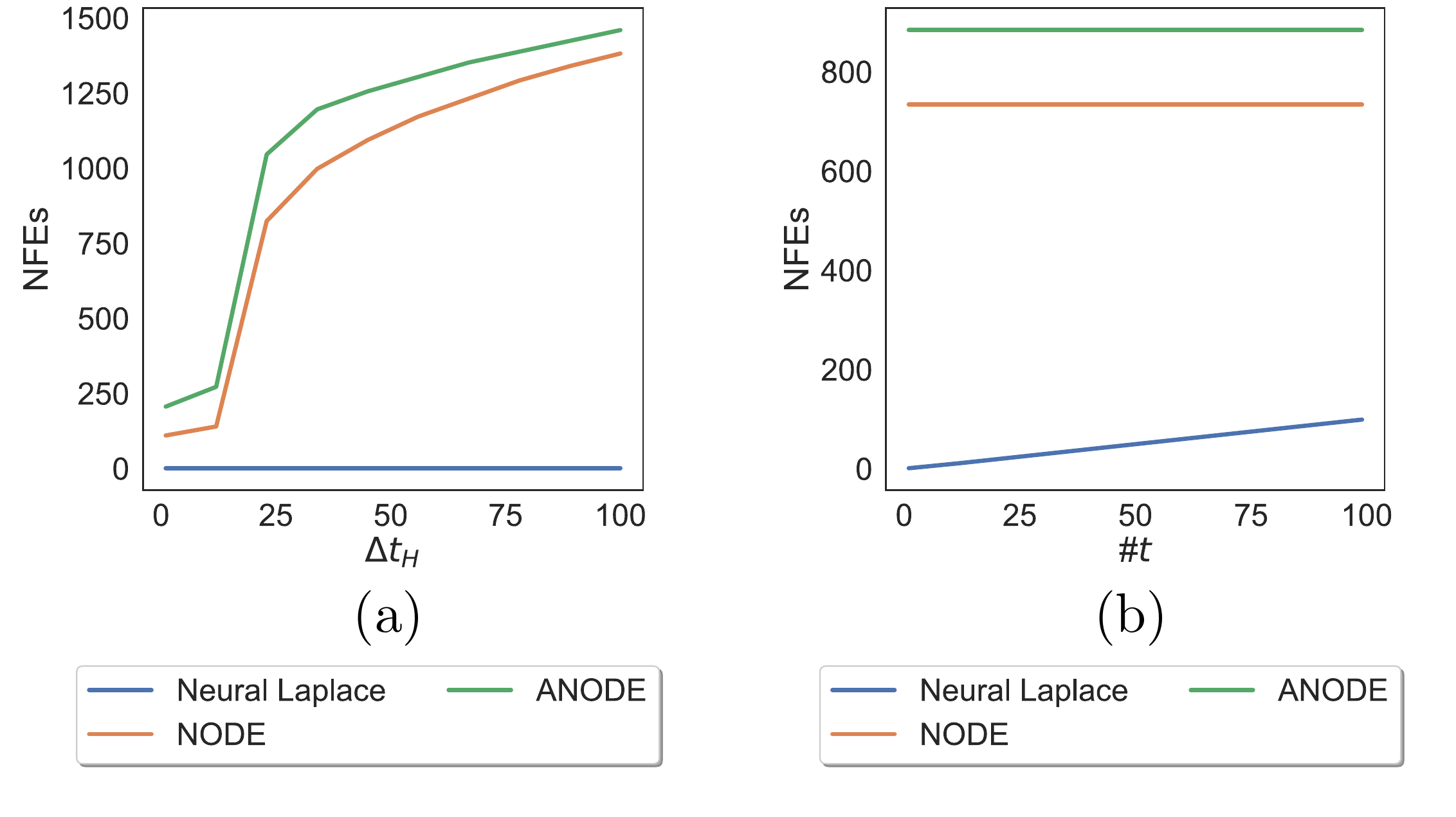}
\caption{NFEs \textbf{(a)} for extrapolating to a single future time point, that is $\Delta t_H$ in the future,
 \textbf{(b)} for increasing the number of time points evaluated within a fixed interval of time.}
\label{nfes_analysis}
\vskip -0.15in
\end{figure}

\textbf{Linear forward evaluations} Extrapolating to a single time point at any future time $t$ only uses a single forward evaluation in Neural Laplace's Laplace representation model $g_\beta \big(\mathbf{p}, u(\mathbf{s}))$, whereas ODE based methods (NODE, ANODE) scale poorly in number of forward evaluations when extrapolating to any increasing future time $t$, as they use numerical stepwise ODE solvers. 
For comparison Figure \ref{nfes_analysis} (a) shows Neural Laplace can use one thousand times less NFEs when extrapolating forwards a $100$ seconds (Appendix \ref{NFEplots}). However Neural Laplace does scale in NFEs linearly with the amount of time points to evaluate at, Figure \ref{nfes_analysis} (b), which is the same for other integral DE methods \citep{DBLP:journals/corr/abs-2110-13040}. Furthermore, Neural Laplace is empirically at least an order magnitude faster to train per epoch than ODE based methods, measuring wall clock (Appendix \ref{wallclocktimes}), and can comparatively converge faster (Appendix \ref{trainlossplots}).
\section{Conclusion and Future work}

We have shown that through a novel geometrical construction, it is possible to learn a useful Laplace representation model for a broad range of DE systems, such as those not able to be modelled by simple ODE based models, those of delay DEs, Stiff DEs, Integro DEs and ODEs with piecewise forcing functions. Neural Laplace can model the same systems ODE based methods can as well, whilst being faster to train and evaluate, using the Inverse Laplace Transform to generate time solutions instead of using costly, DE stepwise numerical solvers.
We hope this work provides a practical framework to learn a Laplace representation of a system, which is immensely useful and popular in the fields of science and engineering \citep{schiff1999laplace}.

In future work we wish to use the learned Laplace representation to investigate the other unique properties of this representation, such as stability analysis, limiting its frequency reconstruction terms, and using the Laplace final limit theorem \citep{schiff1999laplace} as an additional regularizer. 
We note that we chose the stereographic projection because it is the \textit{simplest} and most well-studied bijective map, that maps the entire complex domain to a compact domain \citep{10.5555/26851}.
Future work could include using other complex bijective maps instead.
It could also be interesting to explore other integral transforms with a similar geometrical smoothness map to a compact domain, such as the Mellin transform and others \citep{debnath2014integral} that have unique properties that are advantageous to learn a representation for.

\subsection*{Acknowledgements}

SH would like to acknowledge and thank AstraZeneca for funding.
This work was additionally supported by the Office of Naval Research (ONR)
and the NSF (Grant number: 1722516). 
Additionaly we thank Arsenii Nikolaiev for the helpful discussion.
Furthemore we would like to warmly thank all the anonymous reviewers, alongside research group members of the van der Scaar lab, for their valuable input, comments and suggestions as the paper was developed. All these inputs ultimately improved this paper.

\newpage
\bibliography{main}
\bibliographystyle{icml2022}

\clearpage
\appendix
\section*{Table of supplementary materials}

\begin{enumerate}
    \item Appendix \ref{deapplications}: Applications of DEs
    \item Appendix \ref{ILTaglorithms}: Inverse Laplace Transform Algorithms
    \item Appendix \ref{frequencyfilter}: Limiting Reconstruction Frequencies
    \item Appendix \ref{benchmarkdetails}: Benchmark Implementation and Hyperparameters
    \item Appendix \ref{samplingdes}: Sampling Each DE Dataset
    \item Appendix \ref{capturingstatespacedistribution}: Capturing State Space Distribution
    \item Appendix \ref{NFEplots}: NFE Analysis
    \item Appendix \ref{samplesize}: Sample and Observation Size Scaling
    \item Appendix \ref{dopri5results}: Additional Benchmark Results
    \item Appendix \ref{wallclocktimes}: Benchmark Wall Clock Times
    \item Appendix \ref{trainlossplots}: Training Loss Plots
    \item Appendix \ref{toywaveforms}: Extrapolating Toy Waveforms
    \item Appendix \ref{datasetplots}: Dataset Plots
\end{enumerate}

\textbf{Code}. We have released a PyTorch implementation \citep{paszke2017automatic}, including GPU implementations of several ILT algorithms at \href{https://github.com/samholt/NeuralLaplace}{https://github.com/samholt/NeuralLaplace}. We also have a research group codebase, which can be found at \href{https://github.com/vanderschaarlab/NeuralLaplace}{https://github.com/vanderschaarlab/NeuralLaplace}.

\begin{table*}[!htb]
    \centering
	\caption[]{Applications of the families of DEs captured by Neural Laplace.} 
	\label{Table:DEapplications}
\resizebox{\textwidth}{!}{
\begin{tabular}{l c}
\toprule
Type & Applications                                                       \\ \midrule
ODE   & Dynamical systems \citep{teschl2012ordinary}, Biological systems \citep{su2021deep}                \\
DDE   & Biological systems \citep{moreira2019delay, glass2021nonlinear, verdugo2007delay}, Electrodynamics \citep{lopez2020electrodynamic} \\
IDE   &  Engineering \citep{zill2016differential}, Epidemiology \citep{medlock2004integro}, Finance \citep{sachs2008efficient} \\
Forced ODE & Control Theory \citep{filippov2013differential}, Engineering \citep{boyce2017elementary}  \\
Stiff ODE & Engineering \citep{van1927frequency}, Chemistry \citep{sandu1997benchmarking}   \\\bottomrule
\end{tabular}
}
\end{table*}

\begin{table*}[htb]
	\centering
	\caption[]{Comparison of ILT algorithms that we considered and implemented.}
	\label{Table:ILTComparisonStructure}
\resizebox{\textwidth}{!}{
\begin{tabular}[b]{cccccccc}
\toprule
ILT & Limitations on & $\mathbf{F}(\mathbf{s})$ & Robust & Model & Model & Supports \\ 
& $\mathbf{x}(t)$ & & & sinusoids & exponentials & batching \\ \midrule
Fourier Series Inverse & None    & Complex  & \cmark& \cmark & \cmark   & \cmark     \\
CME & None    & Complex  & \cmark & \cmark & \cmark & \cmark \\
de Hoog& None    & Complex  & \xmark & \cmark & \cmark    & \xmark     \\
Fixed Tablot & No medium     & Complex  & \cmark& \xmark & \cmark    & \cmark     \\ 
& / large frequencies     & & & & & & \\
Stehfest     & No oscillations, & Real part only & \xmark & \xmark & \cmark    & \cmark    \\
& no discontinuities in $\mathbf{x}(t)$ & & & & & & \\
\bottomrule
\end{tabular}
}
    \hfill
\end{table*}

\section{Applications of DEs}
\label{deapplications}

The families of DEs that Neural Laplace can capture in Table \ref{tab:DEs} have wide applications with some of their applications listed in Table \ref{Table:DEapplications}.

\section{Inverse Laplace Transform Algorithms} \label{ILTaglorithms}

We use the well-known ILT Fourier series inverse algorithm (ILT-FSI), as it can obtain the most general time solutions whilst remaining numerically stable \citep{10.1145/321439.321446, deHoog:1982:IMN, kuhlman2012}. Others \citep{kuhlman2012} have shown empirically in a review of ILT methods, ILT-FSI methods are the most robust, although we do not get the same convergence guarantees as with other well known inverse Laplace transforms, such as Tablot's method. However these cannot represent frequencies, i.e. poles of the system that pass its restrictive deformed integral contour of Eq. \ref{ilt}, leading to only sufficient representation of solutions of decaying exponentials.

We implemented five inverse Laplace transform algorithms, choosing them for their good performance, somewhat ease of implementation and robustness as indicated in the review of \citep{kuhlman2012}. Implemented in PyTorch \citep{paszke2017automatic}. They are, Fourier Series Inverse \citep{10.1145/321439.321446, deHoog:1982:IMN}, de Hoog \citep{deHoog:1982:IMN}, Fixed Tablot \citep{Abate2004MultiprecisionLT}, Stehfest \citep{10.1145/361953.361969} and concetrated matrix exponentials (CME) \citep{horvath2020numerical}. We compare them in table \ref{Table:ILTComparisonStructure} and in the following we explain each one, comparing each ILT to determine which one best suits our purpose of modelling arbitrary solutions. For a detailed in depth comparison and description of their properties (excluding CME) see the review of \citep{kuhlman2012}. They are as follows,

\textbf{Fourier Series Inverse} Expands Equation \ref{ilt} into an expanded Fourier transform, approximating it with the trapezoidal rule. This keeps the Bromwich contour parallel to the imaginary axis, and shifts it along the real axis, following the definition in Equation \ref{ILTFourier}, i.e. $\sigma \propto \frac{1}{t}$. It is fairly easy to implement and scale to multiple dimensions. We denote $s=\sigma + i \omega$ and we can express Equation \ref{ilt} as,
\begin{equation} \label{ILTFourier}
\begin{split}
    \mathbf{x}(t) & =  \frac{1}{\pi}e^{\sigma t} \int_0^\infty \Re \left\{ F(s) e^{i \omega t} \right\} d \omega \\
    & \approx \frac{1}{T} e^{\sigma t} \left[ \frac{F(\sigma)}{2}  + \sum_{k=1}^{2N} \Re \left\{  F \left( \sigma + \frac{ik\pi}{T} \right)e^{\frac{ik \pi t}{T}} \right\}      \right]
\end{split}
\end{equation}
Where we approximate the first Fourier ILT, Equation \ref{ILTFourier} as a discretized version, using the trapezoidal rule with step size $\frac{\pi}{T}$ \citep{10.1145/321439.321446} and evaluating $s$ at the approximation points $s_k=\sigma + \frac{ik\pi}{T}$ in the trapezoidal summation. We follow \citep{kuhlman2012} to set the parameters of $\sigma=\alpha-\frac{\log(\epsilon)}{T}$, with $\alpha=$1e-3, $\epsilon=10 \alpha$, and the scaling parameter $T=2t$. This gives the query function,
\begin{equation} \label{FourierQuery}
\begin{split}
s_k(t) & = \text{1e-3} - \frac{\log(\text{1e-2})}{2t} + \frac{ik\pi}{2t} \\
\mathcal{Q}(t) & = [s_0(t), \ldots, s_{2N}(t)]^T
\end{split}
\end{equation}
Where we model the equation with $2N + 1$ reconstruction terms, setting $N=16$ in experiments, and use double point floating precision to increase the numerical precision of the ILT. 

The ILT-FSI, of Equation \ref{ILTFourier} provides guarantees that we can always find the inverse from time $t: 0 \rightarrow \infty$, given that the singularities of the system (i.e. the points at which $F(s) \to \infty$) lie left of the contour of integration, and this puts no constraint on the imaginary frequency components we can model. Of course in practice, we often do not model time at $\infty$ and instead model up to a fixed time in the future, which then bounds the exponentially increasing system trajectories, and their associated system poles that we can model $\sigma \propto \frac{1}{t}$.

\textbf{de Hoog} Is an accelerated version of the Fouier ILT, defined in Equation \ref{ILTFourier}. It uses a non-linear double acceleration, using Padé approximation along with a remainder term for the series \citep{deHoog:1982:IMN}. This is somewhat complicated to implement, due to the recurrence operations to represent the Padé approximation, due to this although higher precision \citep{kuhlman2012}, the gradients have to propagate through many recurrence relation paths, making it slow to use in practice compared to Fourier (FSI), however more accurate when we can afford the additional time complexity.

\textbf{Fixed Tablot} Deforms the Bromwich contour around the negative real axis, where $\mathbf{F}(\mathbf{s})$ must not overflow as $\mathbf{s} \to -\infty$, and makes the Bromwich contour integral rapidly converge as $\mathbf{s} \to -\infty$ causes $e^{\mathbf{s}t} \to 0$ in Equation \ref{ilt}. We implemented the Fixed Tablot method \citep{Abate2004MultiprecisionLT, 10.1093/imamat/23.1.97}, which is simple to implement. However it suffers from not being able to model solutions that have large sinusoidal components and instead is optimized for modelling decaying exponential solutions. We note that whilst it can approximate some small sinusoidal components, for an adaptive time contour as in \citep{kuhlman2012}, the sinusoidal components that can be represented decrease when modelling longer time trajectories, and in the limit for long time horizons, allow only representations of decaying exponentials.

\textbf{Stehfest} Uses a discrete version of the Post-Widder formula \citep{doi:10.1080/10652460108819314} that is an approximation for Equation \ref{ilt} using a power series expansion of real part of $\mathbf{s}$. It has internal terms that alternate in sign and become large as the order of approximation is increased, and suffers from numerical precision issues for large orders of approximation. It is fairly easy to implement.

\textbf{CME} Concentrated matrix exponential (CME), uses a similar form to that of the Fourier Series Inverse, approximating Equation \ref{ilt} with the trapezoidal rule \citep{horvath2020numerical}. This uses the form of,
\begin{align} \label{generalILTform}
      \mathbf{x}(t) \approx \frac{1}{T}\sum_{k=1}^{2N} \eta_k F \left(\frac{\beta_k}{T}\right)
\end{align}
The coefficients $\eta_k, \beta_k$ are determined by a complex procedure, with a numerical optimization step involved \citep{horvath2020numerical}. This provides a good approximation for the reconstruction and the coefficients of up to a pre-specified order can be pre-computed and cached for low complexity run time \citep{horvath2020numerical}.
Similarly to Fourier (FSI), CMEs Bromwich contour remains parallel to the imaginary axis and is shifted along the real axis, i.e. $\sigma \propto \frac{1}{t}$. It is moderately easy to implement when using pre-computed coefficients and scale to multiple dimensions. We use $N=16$, to set the reconstruction terms.

The review author of \citep{kuhlman2012} found that for boundary element simulations the Fourier based, ILT Fourier series algorithms were the most robust and most precise. Our test comparison in Table \ref{ILTtimeres} confirms this, with de Hoog being the most precise, however implementing the recurrence operation in PyTorch, causes it to perform slowly as a decoder due to the significantly more gradient operators and path length compared to that of Fourier series inverse ILT.
All these discussed ILT algorithms are implemented and included in the code for this paper.
\begin{table}[!htb]
\centering
  \caption[]{Inverse Laplace Transform algorithms compared against the numerical inversion $F(s)=\frac{s}{s^2+1}$ to it's analytic time function of $x(t)=\cos(t)$. We evaluate for the times $t\in[0, 10.0]$ for 1,000 linearly spaced time points. We fix each algorithm to use $2N+1$ reconstruction terms per time point, with $N=16$.}
  \label{ILTtimeres}
  \begin{tabular}{l c r}
    \toprule
    ILT & $\text{RMSE}(x(t), \hat{x}(t))$ & Forward pass \\
    Algorithm & & time per $t$ ($\mu s$) \\
    \midrule
    Fourier (FSI) & 0.0171      & 2.6331 \\
    Tablot        & 0.4365      & 3.7520 \\
    Stehfest      & 0.2842      & 0.5839 \\
    de Hoog       & 7.056E-10 & 20.3071 \\    
    CME           & 0.0069      & 1.8759 \\
    \bottomrule
  \end{tabular}
\end{table}

\begin{figure}[!htb]
  \centering
    \includegraphics[width=0.8\textwidth/2]{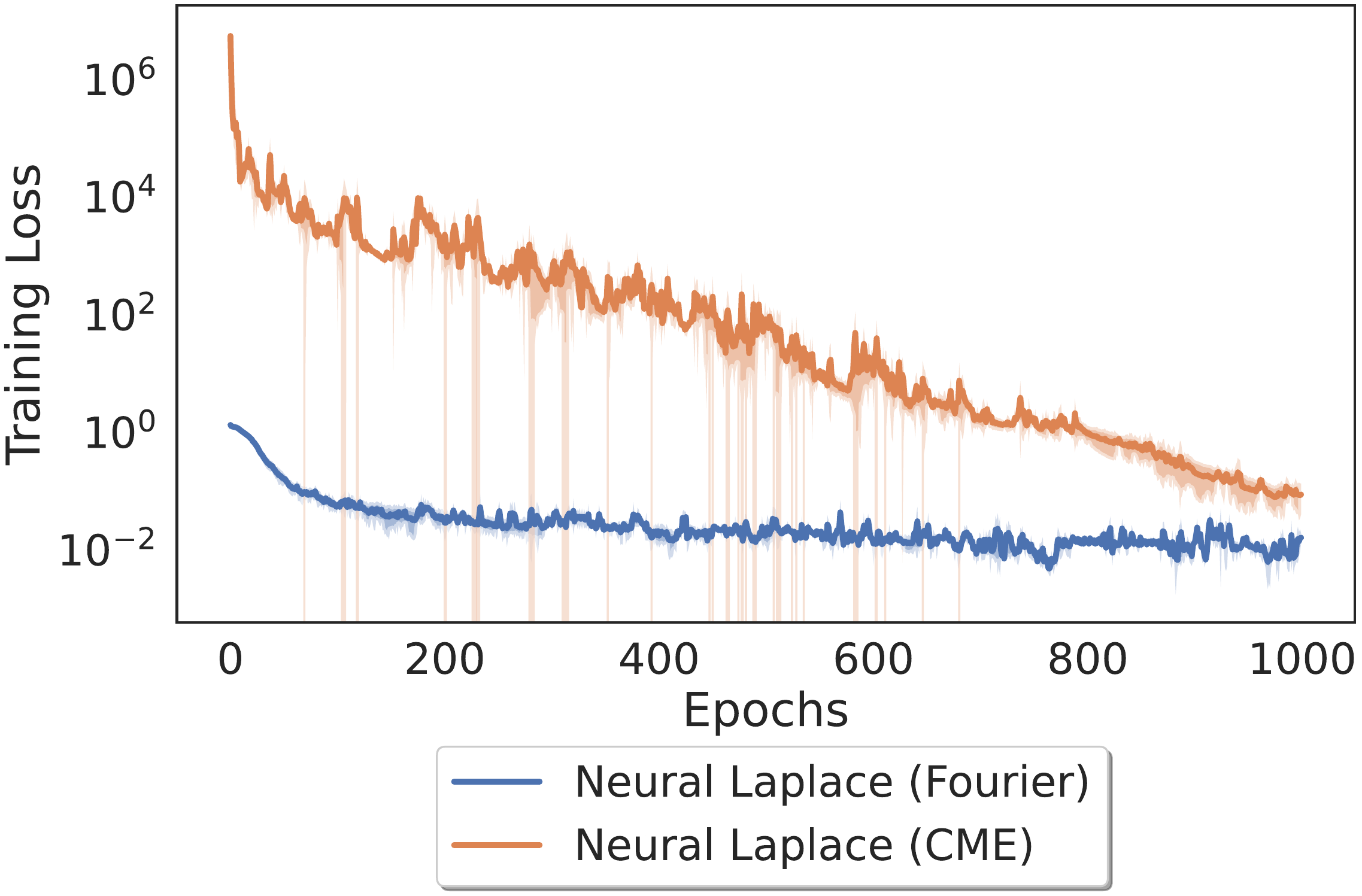}
\caption{Training loss for Lotka-Volterra DDE. Averaged over 3 runs.}
\label{ILTcompar}
\end{figure}

Furthermore Table \ref{ILTtimeres} shows that CME is also competitive compared to Fourier (FSI) ILT algorithm. However we empirically observe in Figure \ref{ILTcompar}, that Neural Laplace converges faster when using the Fourier (FSI) ILT algorithm compared to using the CME ILT algorithm.

\section{Limiting reconstruction frequencies} \label{frequencyfilter}
\begin{figure}[!htb]
    \centering
      \includegraphics[width=0.8\textwidth/2]{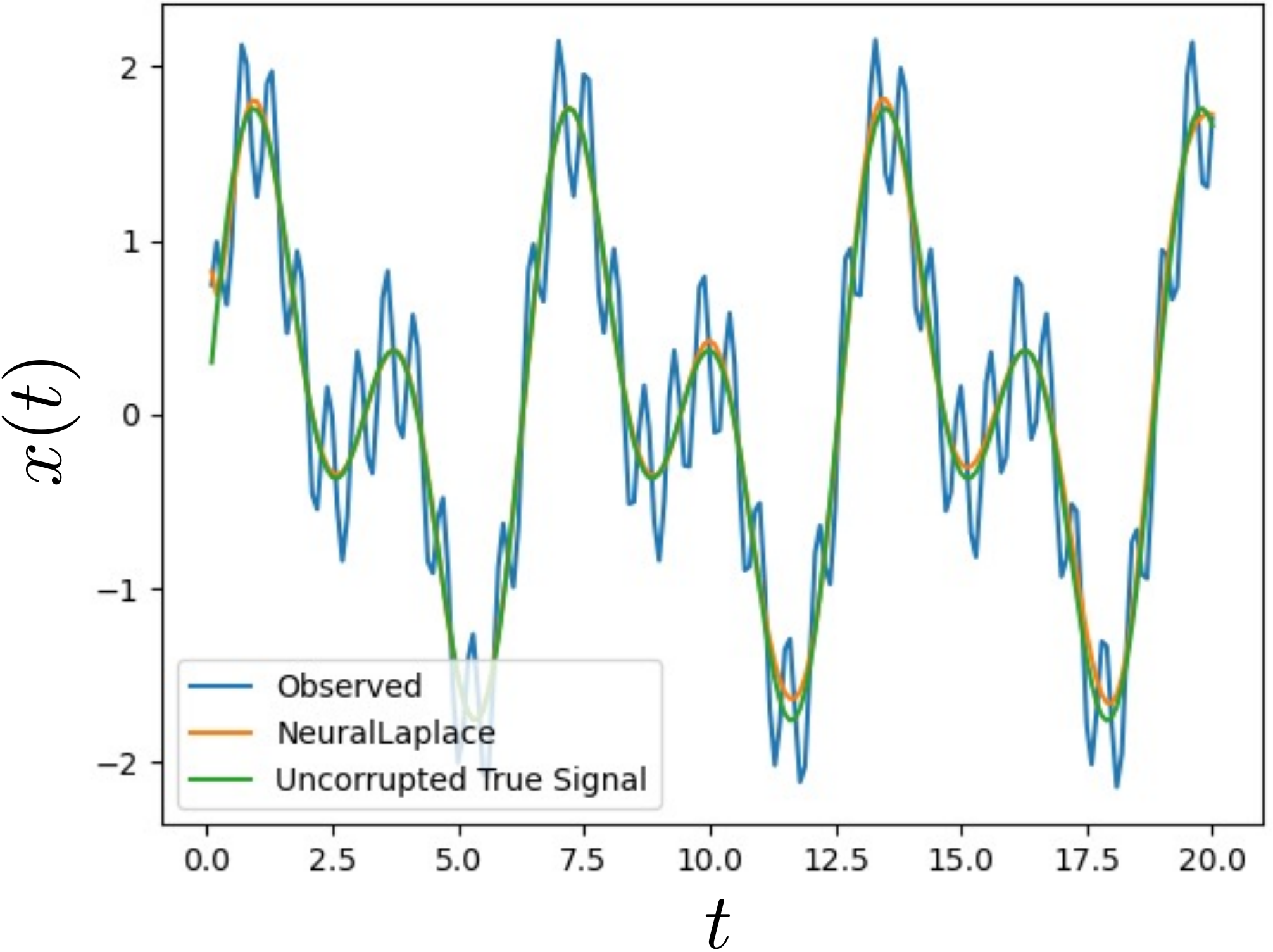}
  \caption{We can limit the frequency that our Neural Laplace model can learn, by applying a low pass frequency filter, to only be able to reconstruct frequencies of interest if we know the frequency range to expect, ideally below the observed noise frequency.}
  \label{freq_filter}
\end{figure}

We consider a toy example of a true signal with high frequency noise as,
\begin{equation}
\begin{split}
y_{\text{uncorrupted}} &= \sin(t) + \sin(2t) \\
y_{\text{corrupted}} &= y_{\text{uncorrupted}} + 0.5 \sin(11t)
\end{split}
\end{equation}
We can explicitly filter our reconstruction frequencies before doing the ILT in Neural Laplace. Here we use a low pass filter, and only allow reconstruction of frequencies below $3$ Hz. We do this by constraining the maximum value of $\phi$ that we can learn, by not allowing any $s$ greater than $|3|$ in the s-domain representation, using Equation \ref{phi0} to set $\phi$. This also limits the exponentials that we can learn as well. Empirically when we do this we can recover the true noise free signal, when training a Neural Laplace model on the corrupted signal. This is advantageous as the true signal is recovered and it was never observed, although we added the prior information that frequencies above a certain threshold are noise and should be disregarded. A plot of this toy example can be seen in Figure \ref{freq_filter}.

\section{Benchmark implementation and hyperparameters} \label{benchmarkdetails}

\begin{table}[!htb]
\centering
  \caption[]{Benchmarks implemented and their number of parameters for each model.}
  \label{benchmarkparams}
  \begin{tabular}{l c }
    \toprule
    Method & \# Parameters \\
    \midrule
NODE	& 17,025 \\
ANODE	 & 17,282 \\
Latent ODE & 	18,565 \\
NF coupling	& 18,307 \\
NF ResNet	& 18,307 \\
Neural Laplace & 17,194 \\
    \bottomrule
  \end{tabular}
\end{table}

For our benchmarks, we tune all methods to have the same number of approximate parameters, as seen in Table \ref{benchmarkparams}, to ensure fair comparison for any gains in modelling complexity. We also set the latent dimension if one exists for each method to be 2, (although we show in Section \ref{ablationstudy}, that Neural Laplace can benefit with a latent dimension greater than 2). To aid the ILT numerical stability, we train and evaluate all models and all data with double point floating precision, as is recommended when using ILTs \citep{kuhlman2012}. We use the Adam optimizer \cite{kingma2017adam} with a learning rate of 1e-3, and batch size of 128. When training we use early stopping using the validation data set with a patience of 100, training for 1,000 epochs unless otherwise stated. These benchmarks are:

\textbf{Neural ODE} \citep{DBLP:journals/corr/abs-1806-07366}, using their code and implementation provided, setting the ODE function $\mathbf{f}(t, \mathbf{x}(t))$ to a 3 layer Multilayer perceptron (MLP), of 128 units, with $\tanh$ activation functions. As NODE does not have an encoder, we set the initial value to the last observed trajectory value at the last observed time. To allow for fair comparison we use the semi-norm trick for faster back propagation \cite{DBLP:journals/corr/abs-2009-09457}, and use the 'euler' solver unless otherwise stated. Using the reconstruction MSE for training.

\textbf{Augmented Neural ODE} \citep{NEURIPS2019_21be9a4b}, we also use their implementation provided, setting the ODE function $\mathbf{f}(t, \mathbf{x}(t))$ to be a 3 layer MLP, of 128 units, with $\tanh$ activation functions, with an additional augmented dimension of zeros. Again ANODE does not have an encoder, so we set the initial value to the last observed trajectory value at the last observed time, and also use the semi-norm trick, and use the 'euler' solver unless otherwise stated. Additionally we use the reconstruction MSE for training.

\textbf{Latent ODE} \citep{DBLP:journals/corr/abs-1907-03907}, which uses an ODE-RNN encoder and an ODE model decoder. We use their code provided, setting the units to be 40 for the GRU and ODE function $\mathbf{f}(t, \mathbf{x}(t))$ net, with $\tanh$ activation, which uses the 'dopri5' solver. We also use their reconstruction variational loss function for training.

\textbf{Neural Flows} \citep{DBLP:journals/corr/abs-2110-13040}, we use their code provided, with the coupling flow using 31 units, and the ResNet flow using 26 units. Again we use their reconstruction variational loss function for training.

\textbf{Neural Laplace} This paper, uses a GRU encoder $h_\gamma$ \cite{cho2014properties}, with 2 layers, with 21 units, with a linear layer on the final hidden state which outputs the latent initial condition $\mathbf{p}$. For the Laplace representation model, $g_\beta$ we use a 3 layer MLP with 42 units, with $\tanh$ activations. We use $\tanh$ on the output to constrain the output domain to be $(\theta, \phi) \in \mathcal{D} = (-{\pi}, {\pi}) \times (-\frac{\pi}{2}, \frac{\pi}{2})$ for each observation dimension. For a given trajectory we encode it into $\mathbf{p}$ and concatenate with $u(\mathbf{s})$ as input to $g_\beta$, i.e. $g_\beta \big(\mathbf{p}, u(\mathbf{s}))$.

\begin{table*}[!htb]
	\centering
	\caption[]{Test conditional mutual information (CMI) between the predicted extrapolation
	to that of the ground truth extrapolation. Best results are bolded. Averaged over $5$ runs. Higher mutual information score is best.} \label{Table:CMI}
		\begin{tabular}[b]{l c c c}
			\toprule
& Mackey-Glass & Lotka-Volterra & Stiff Van de \\
Method & DDE & DDE & Pol Oscillator DE  \\ \midrule
NODE                    & 0.2899 $\pm$ 0.0478 & 3.2479 $\pm$ 0.0910 & 2.8658 $\pm$	0.0142 \\
ANODE                   & 0.2865 $\pm$ 0.0502 & 3.2398 $\pm$ 0.0560 & 2.9679 $\pm$	0.1022 \\
Latent ODE              & \textbf{1.3736 $\pm$ 0.1246} & 3.3908 $\pm$ 0.0357 & 2.7692 $\pm$	0.1446 \\
NF Coupling             & \textbf{1.5569 $\pm$ 0.1871} & 2.1696 $\pm$ 0.5000 & 2.7183 $\pm$	0.1766 \\
NF ResNet               & \textbf{1.5730 $\pm$ 0.1615} & 3.4854 $\pm$ 0.1551 & \textbf{3.0832 $\pm$	0.1435} \\
Neural Laplace          & \textbf{1.6011 $\pm$ 0.1571} & \textbf{3.7913 $\pm$ 0.0359} & \textbf{3.3224 $\pm$	0.1157} \\
                     \bottomrule
        \end{tabular} 
    \hfill
\end{table*}

\section{Sampling each DE dataset} \label{samplingdes}

We test the benchmarks for extrapolation and split each sampled trajectory into two equal parts $[0,\frac{T}{2}]$, $[\frac{T}{2}, T]$, encoding the first half and predicting the second half. For each dataset equation we sample 1,000 trajectories, each with a different initial condition giving rise to a different and unique trajectory defined by the same differential equation system. We use a train-validation-test split of 80:10:10, and train each model for 1,000 epochs with a learning rate of 1e-3 and unless otherwise specified we sample each system in the interval of $t \in [0, 20]$ for 200 time points linearly. For each sequential experiment for the same method we set a different random seed. We also use the training set to normalize the train, val and test set.

To sample the delay DE systems, we use a delay differential equation solver of \citet{DDESolver}, to sample the Spiral DDE, Lotka-Volterra DDE, and Mackey–Glass DDE data sets. We use 2,000 samples in the DDE solver and then subsampled the generated trajectories to 200 time points in the time interval defined for the dataset. The Mackey-Glass DDE dataset was sampled in the time interval $[0,100]$ and then scaled down to $[0,20]$ for equal comparison with the benchmarks. Trajectories of it can be seen in Figure \ref{mackeylongrange}, with benchmarks trained for 50 epochs. To sample the stiff DEs, that of the Stiff Van der Pol Oscillator we use the stiff DE solver of 'ode15s' \cite{shampine1997matlab}, we sampled for a given initial condition between the times of 0 to 4000, generating 200 time points for each trajectory. Once the trajectories were generated we reduced the time interval by dividing by 200, to get trajectories from times from 0 to 20, to make it more comparable to other data sets analyzed.

For sampling the Spiral DDE, we set $\mathbf{A}$ to
\begin{equation} \label{spiralddeconsta}
\begin{split}
    \mathbf{A} & = \begin{bmatrix}
-1 & 1 \\
-1 & -1 
\end{bmatrix} \\
\end{split}
\end{equation}

For sampling the Integro DE, we use the analytical general solution for a given initial value (solvable with the Laplace transform method and then inverting back \citep{integrode}). This gives solutions of
\begin{equation} \label{ide_sols_0}
\begin{split}
    x(t)(x(0)) = & \frac{1}{4} \Re \big[e^{(-1-2i)t} \big[(2x(0) + \\ 
    & (x(0) - 1)i)e^{4it}+ \\
    & (2x(0)-(x(0)-1)i)\big]\big] \\
\end{split}
\end{equation}
for a given initial value $x(0)$.

We similarly sampled the ODE with piecewise forcing function using its analytical general solution (which can be generated by the Laplace transform method). This gives solutions of
\begin{equation} \label{ide_sols_1}
\begin{split}
    x(t)(x(0)) = & x(0) \cot(2t) + \\
    & \frac{1}{5} u(t-5) \frac{1}{4} ((t-5) - \frac{1}{2} \sin(2 (t-5))) + \\
    & \frac{1}{5} u(t-10) \frac{1}{4} ((t-5)-(t-10)- \\
    & \frac{1}{2} \sin(2(t-5)) + \frac{1}{2} \sin(2 (t-10)) \\
\end{split}
\end{equation}

for a given initial value $x(0)$ and where $u(t)$ is the Heaviside step function.

\section{Capturing state space distribution} \label{capturingstatespacedistribution}

To investigate whether we capture the state space distribution, we empirically measure the mutual information between the ground truth expected extrapolation and the predicted extrapolation. We report the conditional mutual information (CMI) as each extrapolation depends on the initial condition for the DE system, therefore we condition on the initial condition, Table \ref{Table:CMI}. To compute the conditional mutual information we used the non-parametric entropy estimator toolbox of \citet{ver2000non}. Table \ref{Table:CMI} shows that Neural Laplace is able to capture the state space distribution correctly, that is the extrapolation distribution conditioned on the input observed (initial history) trajectory.

\begin{table*}[!htb]
	\centering
	\caption[]{Test RMSE for Lotka-Volterra DE dataset analyzed. Averaged over 3 runs. We vary the dataset trajectory size $N$ in total from 1,000 trajectories to 30. Best results are bolded. Training for 200 epochs per dataset.} \label{Table:datascaling}
	\resizebox{\textwidth}{!}{
		\begin{tabular}[b]{l c c c c c c}
			\toprule
     $N$           &       1000          &         500         &          250        &         125         &       62            &         30          \\ \midrule
NODE            & 0.4958 $\pm$ 0.0707 & 0.6519 $\pm$ 0.0280 & 0.6757 $\pm$ 0.0818 & 0.9122 $\pm$ 0.0211 & \textbf{0.9196 $\pm$ 0.6280} & \textbf{0.9663 $\pm$ 0.3946} \\
ANODE           & 0.4852 $\pm$ 0.0127 & 0.6499 $\pm$ 0.0863 & 0.6816 $\pm$ 0.0404 & 0.9387 $\pm$ 0.0547 & \textbf{0.8747 $\pm$ 0.5582} & \textbf{0.9682 $\pm$ 0.3370} \\
Latent ODE      & 0.4826 $\pm$ 0.0343 & 0.6629 $\pm$ 0.0349 & 0.6839 $\pm$ 0.0497 & 0.9948 $\pm$ 0.1742 & \textbf{0.9550 $\pm$ 0.4901} & \textbf{0.9547 $\pm$ 0.3125} \\
NF coupling     & 0.8407 $\pm$ 0.0272 & 0.9336 $\pm$ 0.0795 & 0.9233 $\pm$ 0.0834 & 1.1865 $\pm$ 0.0363 & \textbf{1.0881 $\pm$ 0.4945} & \textbf{1.0427 $\pm$ 0.3416} \\
NF ResNet       & 0.4646 $\pm$ 0.1059 & 0.8297 $\pm$ 0.1302 & 0.8069 $\pm$ 0.0989 & 1.1484 $\pm$ 0.1629 & \textbf{0.9640 $\pm$ 0.4599} & \textbf{0.9395 $\pm$ 0.2326} \\
Neural Laplace  & \textbf{0.1371 $\pm$ 0.0330} & \textbf{0.2581 $\pm$ 0.0255} & \textbf{0.3486 $\pm$ 0.0275} & \textbf{0.5716 $\pm$ 0.0840} & \textbf{0.7715 $\pm$ 0.6275} & \textbf{0.9220 $\pm$ 0.3086} \\
                     \bottomrule
        \end{tabular} 
        }
    \hfill
\end{table*}

\begin{table*}[!htb]
	\centering
	\caption[]{Test RMSE for datasets analyzed, adding Gaussian noise, $\mathcal{N}(0,0.1)$ to each trajectory sampled. Best results are bolded. Averaged over $3$ runs.} \label{Table:AdditionalNoiseResult}
		\begin{tabular}[b]{c c c c}
			\toprule
   & Lotka-Volterra      & Integro                 & ODE piecewise         \\
Method     & DDE                 & DE                      & forcing function    \\ \midrule
NODE                    &  .6043 $\pm$	.1126           & .1217 $\pm$ .0020             & 0.2641 $\pm$ 0.0073 \\
ANODE                   &  .5952 $\pm$	.1085           & .1191 $\pm$ .0027             & 0.1642 $\pm$ 0.0035 \\
Latent ODE              &  .2426 $\pm$	.0473           & \textbf{.1002 $\pm$ .0005}             & 0.1542 $\pm$ 0.0011 \\
NF Coupling             &  .6994 $\pm$	.1210           & \textbf{.0999 $\pm$ .0006}             & 0.1242 $\pm$ 0.0016 \\
NF ResNet               &  .2464 $\pm$	.0521           & \textbf{.0998 $\pm$ .0004}             & 0.1058 $\pm$ 0.0021 \\
Neural Laplace          &  \textbf{.1328 $\pm$	.0228}	& \textbf{.0996 $\pm$ .0004}    & \textbf{0.1006 $\pm$ 0.0004} \\
                     \bottomrule
        \end{tabular} 
    \hfill
\end{table*}

\section{NFE Analysis} \label{NFEplots}

Once trained, Neural Laplace can reconstruct anytime with one forward evaluation. We investigated this by training, Neural Laplace, NODE and ANODE on the ODE with piecewise forcing function dataset, using 'dopri5' in the ODE numerical solvers. With the trained models we evaluated them for how many NFEs they use to, (a) extrapolate forwards an increase of $\Delta t_H$ time from the current last time observed, $t=10$, and (b) extrapolate $N$ time points from the last time observed $t=10$ up to the fixed time horizon $t_H=20$. Observing this in Figure \ref{nfes_analysis}, we see that we can extrapolate any time $t$ with one forward pass, whereas NODE methods scale very poorly for long time extrapolation, here achieving a thousand times less NFEs for extrapolating forwards $100$ seconds. We also observe that Neural Laplace does scale linearly in NFEs with the number of time points to evaluate, which is the same for other integral DE methods \cite{DBLP:journals/corr/abs-2110-13040}.

\begin{table*}[!htb]
	\centering
	\caption[]{Test RMSE for toy waveforms analyzed. Best results are bolded. Averaged over $5$ runs.} \label{Table:toywaveforms}
		\begin{tabular}[b]{c c c c}
			\toprule
Method & Sine & Square & Sawtooth \\ \midrule
NODE            & 0.9657 $\pm$ 0.0046 & 0.9769 $\pm$ 0.0056 & 0.9772 $\pm$ 0.0109 \\
ANODE           & 0.7430 $\pm$ 0.0632 & 0.8153 $\pm$ 0.0191 & 0.3001 $\pm$ 0.0152 \\
Latent ODE      & 0.1290 $\pm$ 0.2378 & 0.3443 $\pm$ 0.0973 & 0.3404 $\pm$ 0.1016 \\
NF Coupling     & 0.1060 $\pm$ 0.0535 & 0.2768 $\pm$ 0.0340 & 0.4443 $\pm$ 0.0662 \\
NF ResNet       & 0.1482 $\pm$ 0.0712 & 0.2176 $\pm$ 0.0177 & 0.3790 $\pm$ 0.0645 \\
Neural Laplace  & \textbf{0.0063 $\pm$ 0.0010} & \textbf{0.1678 $\pm$ 0.0067} & \textbf{0.1600 $\pm$ 0.0179} \\
                     \bottomrule
        \end{tabular} 
    \hfill
\end{table*}

\begin{table}[!htb]
\centering
  \caption[]{Average wall clock time taken to train on one epoch, 1,000 trajectories on the Lotka-Volterra DDE dataset.}
  \label{wallclocktime}
  \begin{tabular}{l r}
    \toprule
    Method & seconds per epoch \\
    \midrule
NODE (dopri5)	& 13.78 \\
NODE (euler)  	& 1.92 \\
ANODE (dopri5)	& 17.23 \\
ANODE (euler)	& 4.56 \\
Latent ODE      &         	3.56 \\
NF Coupling     &         	10.83 \\
NF ResNet       &         	0.1 \\
Neural Laplace  &         	0.1 \\
    \bottomrule
  \end{tabular}
\end{table}

\section{Sample and observation size scaling} \label{samplesize}

\textbf{Sample size scaling.} We also investigated how we compare to the benchmarks with varying the number of trajectories in a dataset. We see in Table \ref{Table:datascaling}, when varying the dataset trajectory size $N$, from 1,000 down to 30 (with each trajectory consisting of 200 time points) on the Lotka-Volterra dataset, where we trained each dataset for 200 epochs. We observe that Neural Laplace is able to remain competitive down to 125 trajectories in a dataset compared to the benchmarks, however with trajectories lower than 125, all benchmarks compare the same and this continues for lower trajectory sizes. As expected with smaller numbers of trajectories in a dataset all methods suffer from increased error (increasing RMSE), as they have less data to train on.

\begin{figure}[!htb]
  \centering
    \includegraphics[width=0.8\textwidth/2]{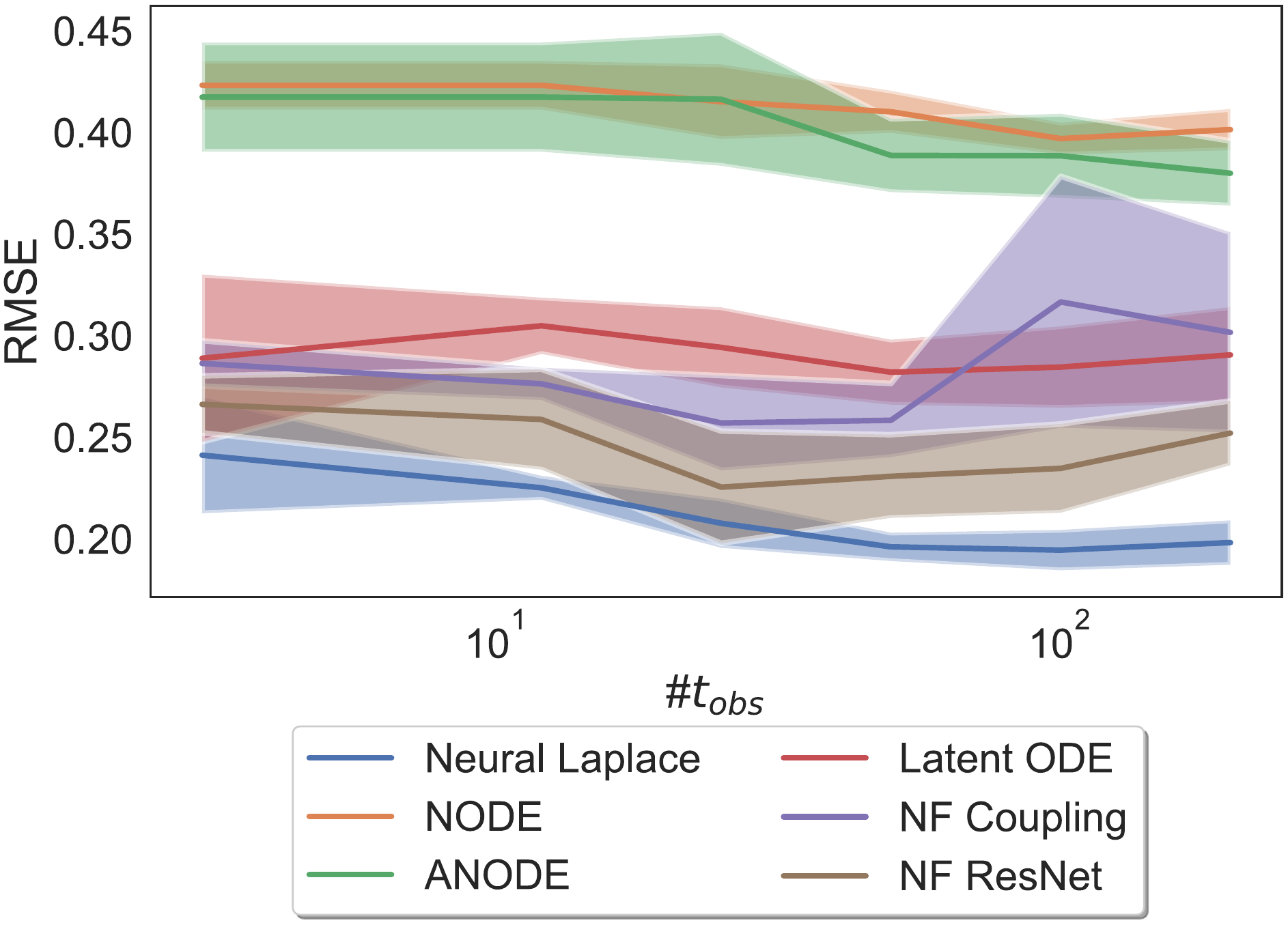}
\caption{Test RMSE for Stiff Van de Pol Oscillator DE dataset analyzed. Averaged over 5 runs. We vary the number of observed points \#$t_{obs}$ for each trajectory. Training for 200 epochs per dataset.}
\label{ObservationSizeScaling}
\end{figure}

\textbf{Observation size scaling}. We further varied the number of observed points for the same extrapolation points on the Stiff Van de Pol Oscillator DE dataset, shown in Figure \ref{ObservationSizeScaling}.
Each dataset was trained for 200 epochs, with 1000 sampled trajectories each with Gaussian noise, $\mathcal{N}(0,0.01)$.
Neural Laplace consistently outperforms the benchmarks, indicating its robustness to the number of observed points. 

\section{Additional Benchmark Results} \label{dopri5results}
For fair comparison of the ODE based benchmarks, we also ran NODE and ANODE with the flexible solver method of 'dopri5'. Table \ref{Table:AdditionalMSE} shows the test RMSE results, we observe Neural Laplace remains competitive.
\begin{table}[!htb]
	\centering
	\caption[]{Test RMSE for datasets analyzed. Best result is bolded. Averaged over $5$ runs.} \label{Table:AdditionalMSE}
		\begin{tabular}[b]{c c c}
			\toprule
& Lotka-Volterra & ODE piecewise \\
Method & DDE &  forcing function \\ \midrule
NODE                   & 0.5880 $\pm$ 0.0398 & 0.1945 $\pm$ 0.0134 \\
ANODE                  & 0.5673 $\pm$ 0.0744 & 0.0769 $\pm$ 0.0039 \\
Neural Laplace         & \textbf{0.0427 $\pm$	0.0066} & \textbf{0.0037	$\pm$ 0.0004} \\
                     \bottomrule
        \end{tabular} 
    \hfill
\end{table}

\begin{figure*}[!htb]
    \centering
  \includegraphics[width=\textwidth]{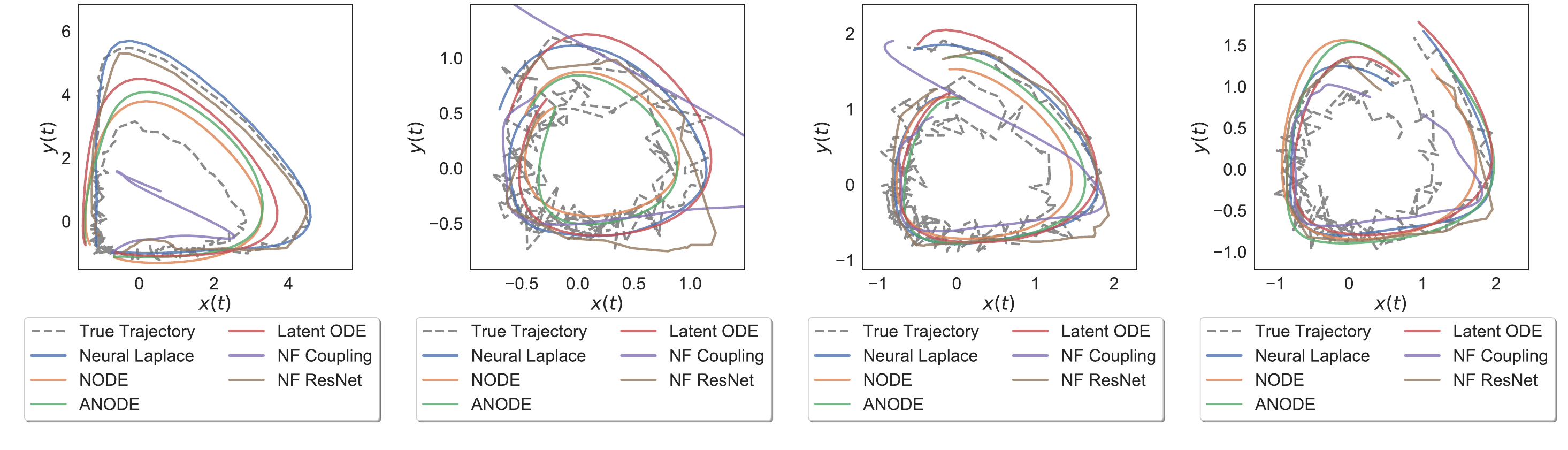}
  \caption{Lotka-Volterra DDE randomly sampled test state plots, with Gaussian noise added all trajectories of $\mathcal{N}(0,\epsilon)$, with $\epsilon=0.1$.}
  \label{noisy_volt_trajs}
\end{figure*}

We also observe the same pattern seen in Table \ref{Table:MainMSE}, when we add noise to the  sampled DE systems. Table \ref{Table:AdditionalNoiseResult} shows results for Gaussian noise added to all the sampled trajectories of $\mathcal{N}(0,0.01)$, and Figure \ref{noisy_volt_trajs} shows some sampled test trajectories on one of these (Lotka-Volterra DDE) noise corrupted datasets.

\section{Benchmark Wall Clock Times} \label{wallclocktimes}

We measured the time to train on one epoch of 1,000 trajectories for each benchmark tested, detailed in Table \ref{wallclocktime}, averaged over training for a 1,000 epochs. For completeness we include 'euler' and 'dopri5' solvers for NODE and ANODE methods. We observe that Neural Laplace is at least one order magnitude faster compared to ODE based solver methods, and in some cases up two orders of magnitude faster. We trained and took these readings on a Intel Xeon CPU @ 2.30GHz, 64GB RAM with a Nvidia Tesla V100 GPU 16GB.

\section{Training Loss Plots} \label{trainlossplots}

\begin{figure}[!htb]
    \centering
    \includegraphics[width=0.95\textwidth/2]{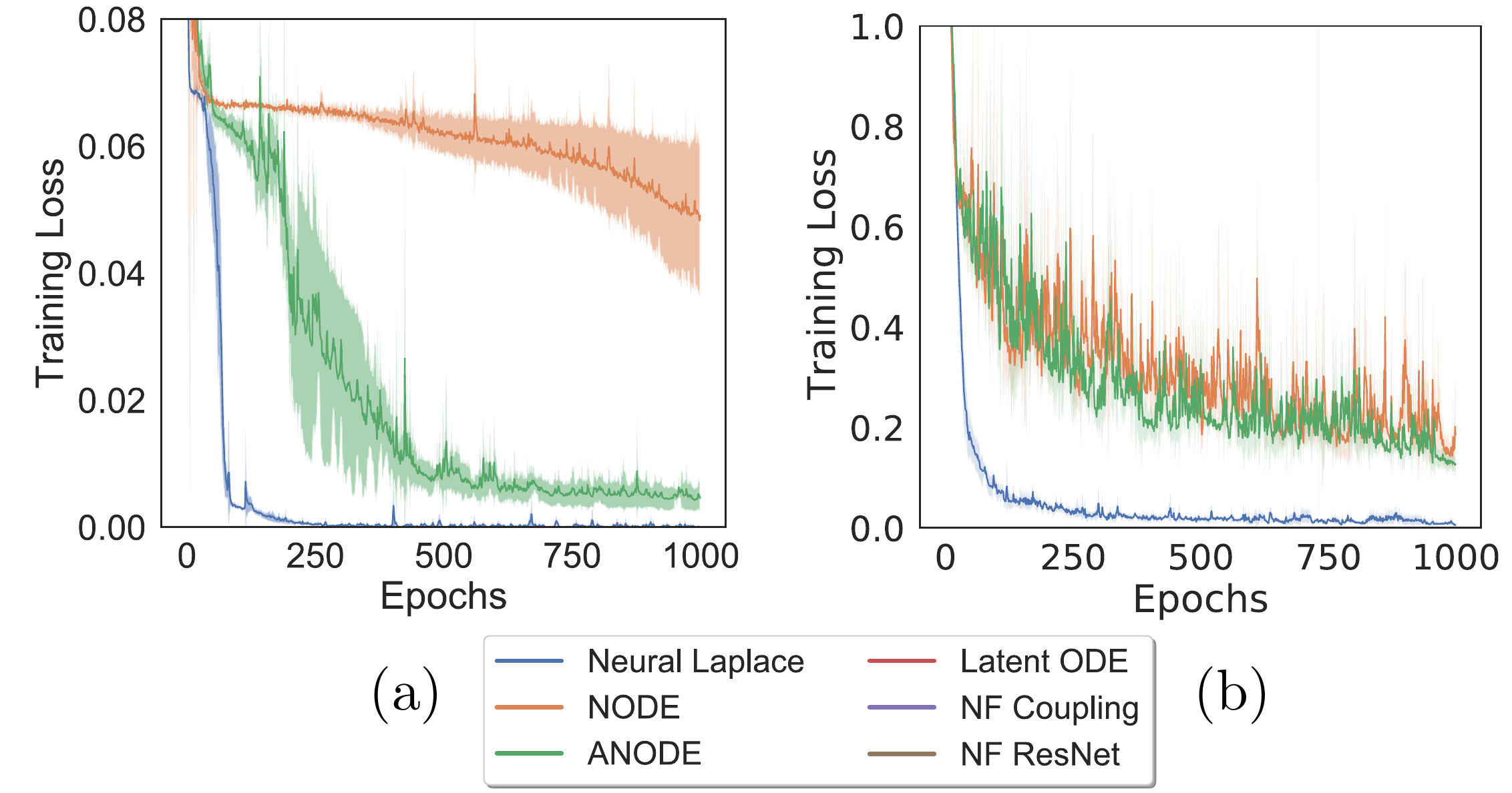}
  \caption{Training loss versus epochs, averaged over 5 runs, with standard deviation bars, training on the (a) ODE with piecewise forcing function dataset, (b) Lotka-Volterra DDE dataset.}
  \label{train_lossplots}
\end{figure}

Training loss plots against epochs can be seen in Figure \ref{train_lossplots}. Empirically we see Neural Laplace can converge faster than the other benchmark methods. 

\section{Extrapolating Toy Waveforms} \label{toywaveforms}

\begin{figure*}[!htb]
    \centering
  \includegraphics[width=\textwidth]{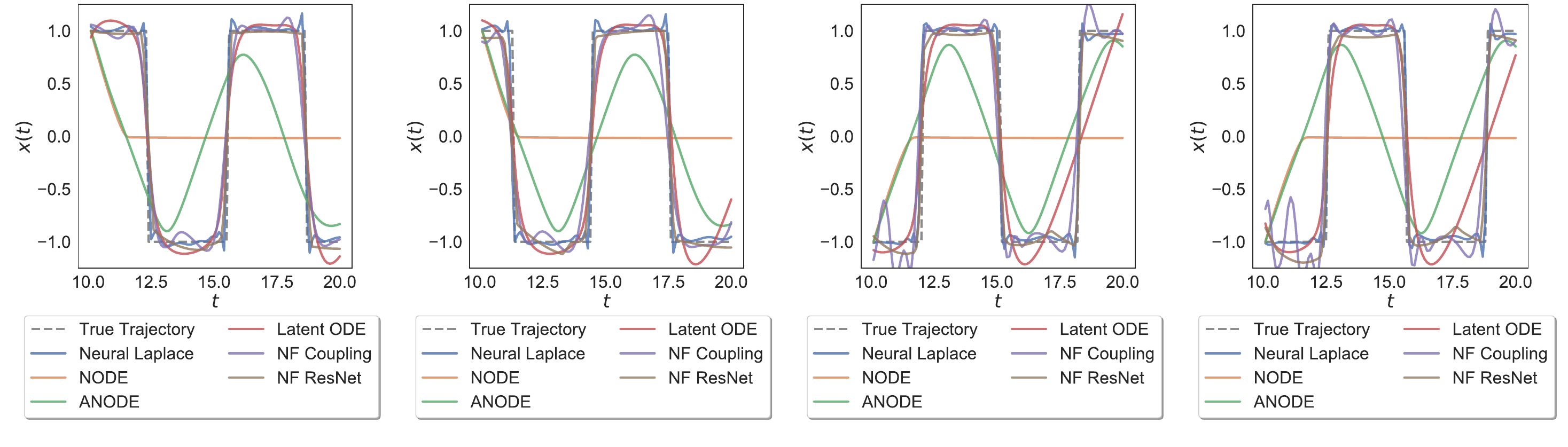}
  \caption{Square waveform randomly sampled test trajectory plots.}
  \label{square_trajs}
\end{figure*}

\begin{figure*}[!htb]
    \centering
  \includegraphics[width=\textwidth]{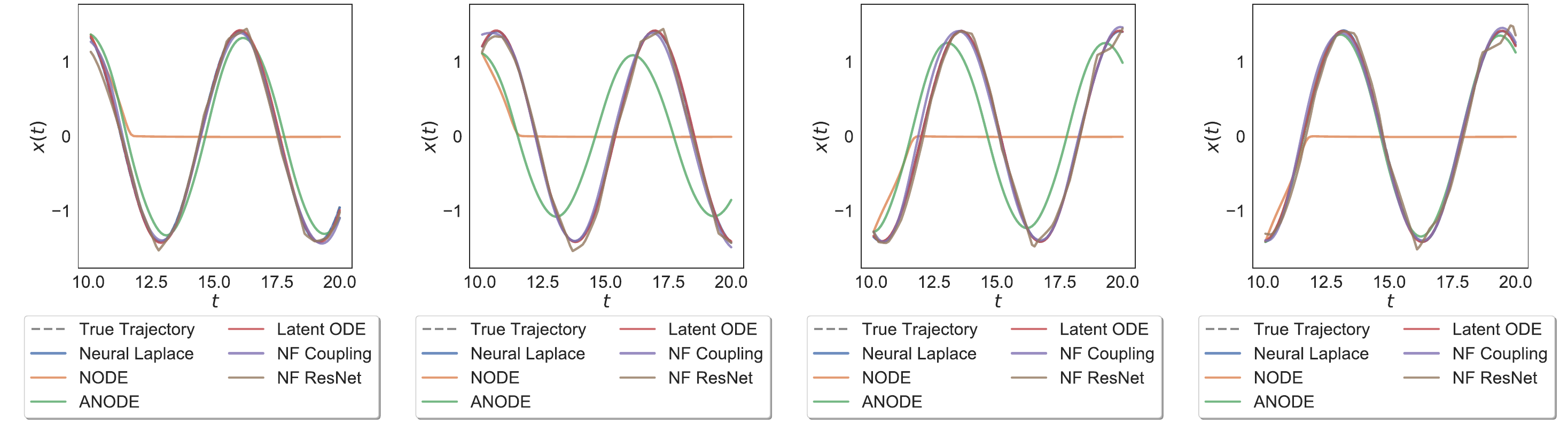}
  \caption{Sine waveform randomly sampled test trajectory plots.}
  \label{sine_trajs}
\end{figure*}

\begin{figure*}[!htb]
    \centering
  \includegraphics[width=\textwidth]{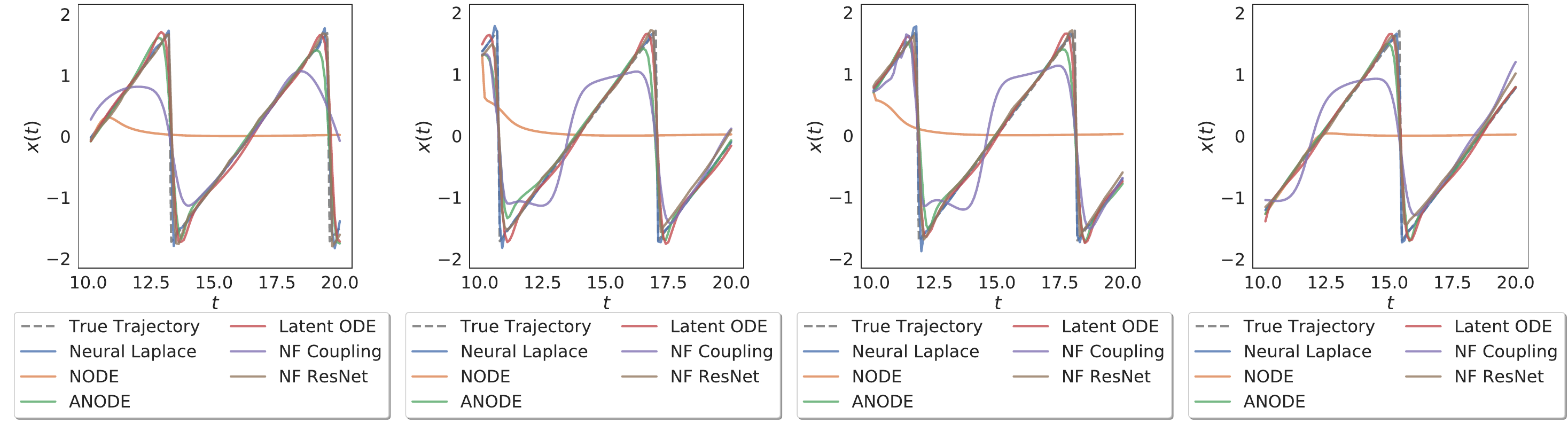}
  \caption{Sawtooth waveform randomly sampled test trajectory plots.}
  \label{sawtooth_trajs}
\end{figure*}

We also explore the benchmarks and Neural Laplace at extrapolating toy waveform signals, that of a sawtooth, square and sine waveform. These are interesting to extrapolate as they are periodic and some contain discontinuities (square and sawtooth). We sampled each from the interval of $t\in[0,20]$, with a period of $2\pi$ for each waveform. We sampled different initial values by sampling a translation from $[0,2\pi]$ to generate different trajectories. These are given as \textit{sawtooth} $x(t)=\frac{t}{2 \pi} - \text{floor}(\frac{t}{2 \pi})$, \textit{square} $x(t)=2 (1-\text{floor}\left(\frac{t}{\pi}\right) \% 2)$ and \textit{sine} $x(t)=\sin(t)$. The results of the methods in extrapolating these waveforms can be seen in Table \ref{toywaveforms}, with illustrations in Figures \ref{square_trajs}, \ref{sine_trajs}, \ref{sawtooth_trajs}.

\section{Dataset Plots} \label{datasetplots}

\begin{figure*}[!htb]
    \centering
  \includegraphics[width=\textwidth]{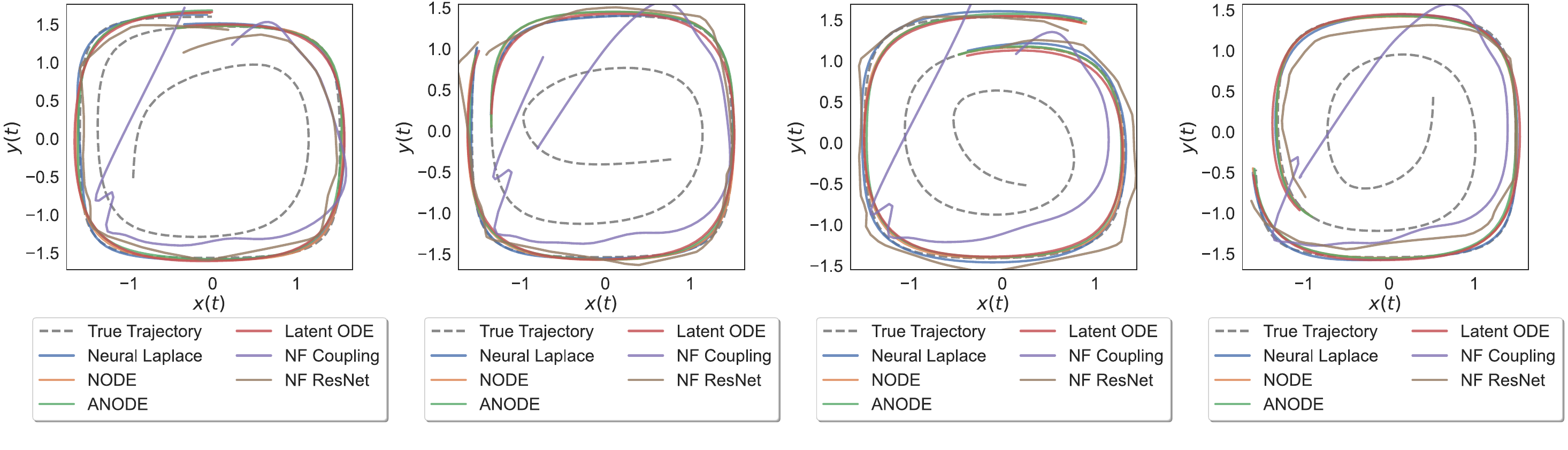}
  \caption{Spiral DDE randomly sampled test state plots.}
  \label{dataset_spiraldde_trajs}
\end{figure*}

\begin{figure*}[!htb]
    \centering
  \includegraphics[width=\textwidth]{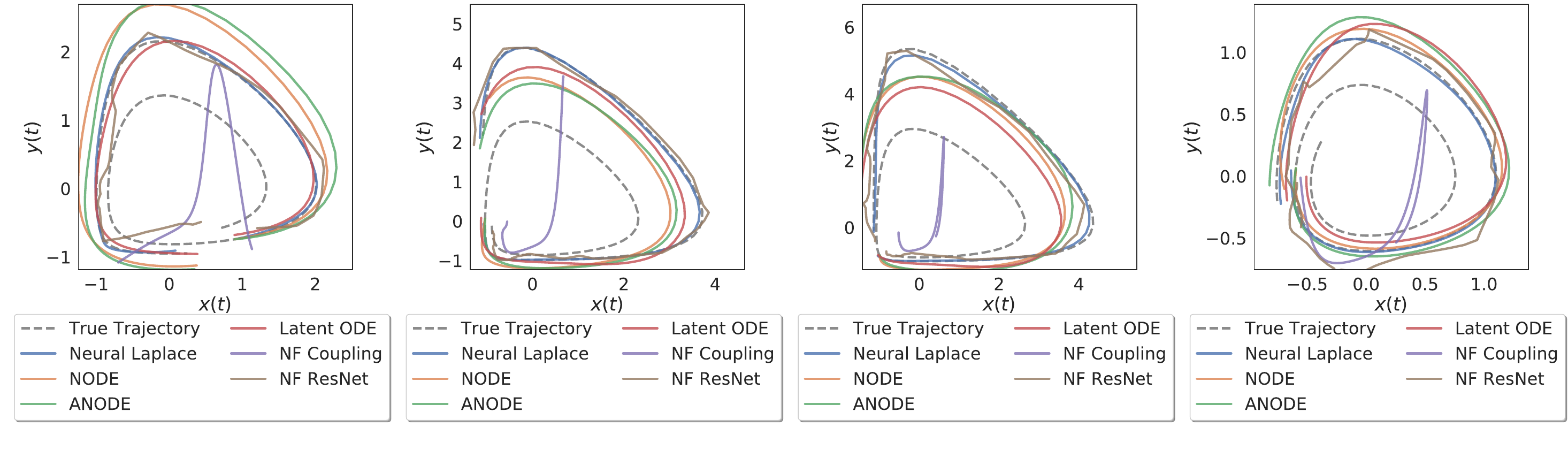}
  \caption{Lotka-Volterra DDE randomly sampled test state plots.}
  \label{dataset_lotka_trajs}
\end{figure*}

\begin{figure*}[!htb]
    \centering
  \includegraphics[width=\textwidth]{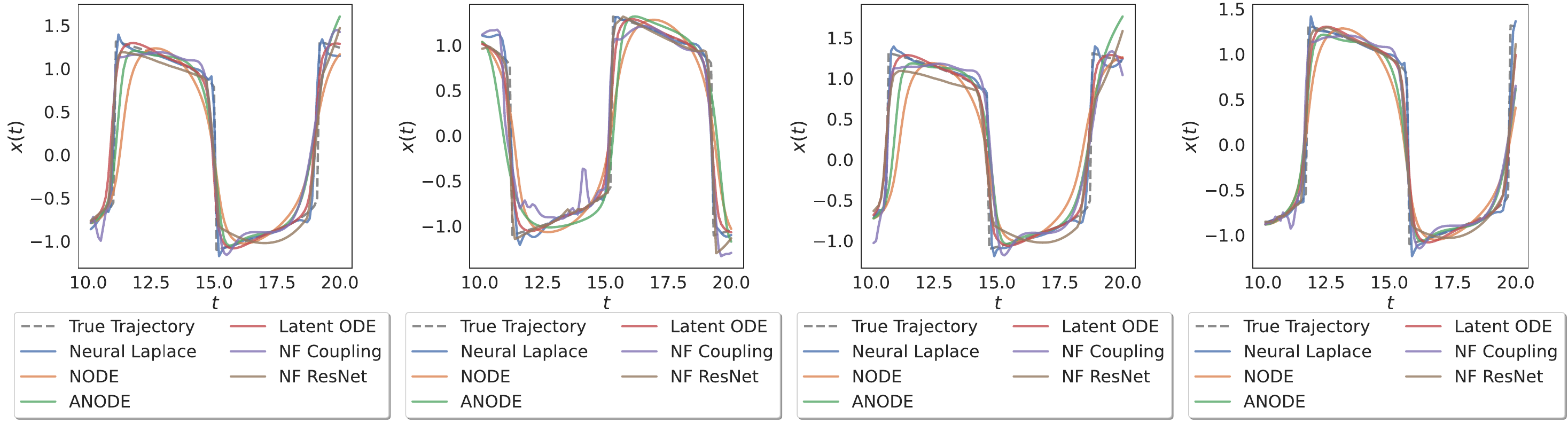}
  \caption{Stiff Van der Pol Oscillator DE randomly sampled test trajectory plots.}
  \label{dataset_stiff_trajs}
\end{figure*}

\begin{figure*}[!htb]
    \centering
  \includegraphics[width=\textwidth]{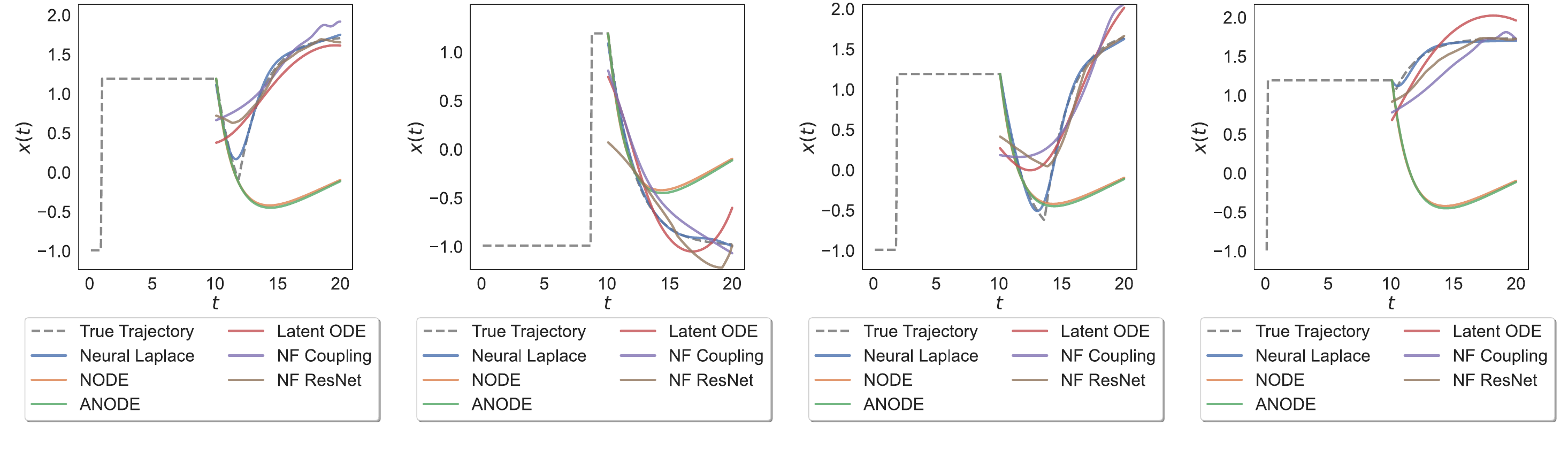}
  \caption{Mackey–Glass DDE modified to exhibit long range dependencies randomly sampled test trajectory plots.}
  \label{mackey_trajs}
\end{figure*}

\end{document}